%% file: LeTO.tex
\documentclass[doublecolumn]{IEEEtran}

\include{before_document}

\newcommand{\red}{\textcolor{red}}
\usepackage{xcolor}
\newcommand{\green}[1]{\textcolor[rgb]{0,0.5,0}{#1}}
\usepackage{graphicx}
\usepackage{caption}
\usepackage{verbatim}

\begin{document}
\title{LeTO: Learning Constrained Visuomotor Policy with Differentiable Trajectory Optimization}
\author{
Zhengtong Xu, Yu She$^{*}$
\thanks{$^{*}$Address all correspondence to this author.}
\thanks{Zhengtong Xu and Yu She are with the Edwardson School of Industrial Engineering, Purdue University, West Lafayette, USA  (E-mail: \{xu1703, shey\}@purdue.edu).}
}

\maketitle

\begin{abstract}
This paper introduces LeTO, a method for learning constrained visuomotor policy with differentiable trajectory optimization. Our approach integrates a differentiable optimization layer into the neural network. By formulating the optimization layer as a trajectory optimization problem, we enable the model to end-to-end generate actions in a safe and constraint-controlled fashion without extra modules. Our method allows for the introduction of constraint information during the training process, thereby balancing the training objectives of satisfying constraints, smoothing the trajectories, and minimizing errors with demonstrations. This ``gray box" method marries optimization-based safety and interpretability with powerful representational abilities of neural networks. We quantitatively evaluate LeTO in simulation and in the real robot. The results demonstrate that LeTO performs well in both simulated and real-world tasks. In addition, it is capable of generating trajectories that are less uncertain, higher quality, and smoother compared to existing imitation learning methods. Therefore, it is shown that LeTO provides a practical example of how to achieve the integration of neural networks with trajectory optimization. We release our code at https://github.com/ZhengtongXu/LeTO.
\end{abstract}
\def\abstractname{Note to Practitioners}
\begin{abstract}
LeTO is driven by the goal of developing an imitation learning algorithm capable of generating safe and constraint-satisfying robotic behaviors. The idea of imitation learning is to enable the robot to learn from human demonstrations of certain tasks. Subsequently, the robot is able to autonomously perform the learned tasks on its own. Thanks to the powerful representational and fitting capabilities of neural networks, imitation learning can let robots perform complex manipulation tasks. However, neural networks often exhibit a certain level of uncertainty and lack theoretical safety guarantees. For robotic systems, it is crucial that robot behaviors meet specific constraints; otherwise, the system may not be sufficiently reliable. Therefore, we introduce LeTO, an approach that integrates trajectory optimization with neural networks to generate actions that not only achieve manipulation tasks, but also comply with constraints. This improves the interpretability, safety, and reliability of robot policies acquired through imitation learning, facilitating their deployment in scenarios with high safety requirements.
\end{abstract}

\section{Introduction}\label{sec:intro}

Imitation learning \cite{pomerleau1988alvinn} focuses on the derivation of robot policies from demonstrations. This process can be formulated as
a supervised learning task, with the aim of learning the mapping
between observations and robot actions. {Through imitation learning, robots can perform highly complex and diverse tasks \cite{chi2024universal, xu2024unit,fu2024mobile,zhu2022viola,wang2024poco,xu2023gan,pignat2021generative}. Moreover, recent studies have shown the feasibility of conducting imitation learning directly from manipulation videos \cite{bahl2022human,wang2023mimicplay,qian2023robot}.}

However, ensuring the safety of robots is very important due to the interaction between robots and the real world. In contrast to optimization-based trajectory generation methods, imitation learning often exhibits greater uncertainty, which can lead to system instability, reduced robustness, and safety concerns. Hence, compared to other supervised learning problems, this poses novel and exceptional challenges in safety for imitation learning.

Existing research focus on different aspects and challenges of imitation learning, such as improving action accuracy by addressing compounding errors \cite{zhao2023learning} and representing multi-modal distributions by implicit policy \cite{jarrett2020strictly, florence2022implicit, chi2023diffusionpolicy}. However, these methods overlook critical aspects of safety and smoothness of the generated trajectories. For robotic systems, their behaviors must meet specific constraints to ensure the safety of the system. Moreover, for certain tasks, robot actions also need to meet specific constraints to ensure their successful completion. In addition, adding explainable constraints for neural network-based algorithms, often called ``black boxes", is much harder than for traditional model-based methods.

\begin{figure}[t]
\centering
\begin{overpic}[trim=0 0 650 0,clip, width=0.5\textwidth]{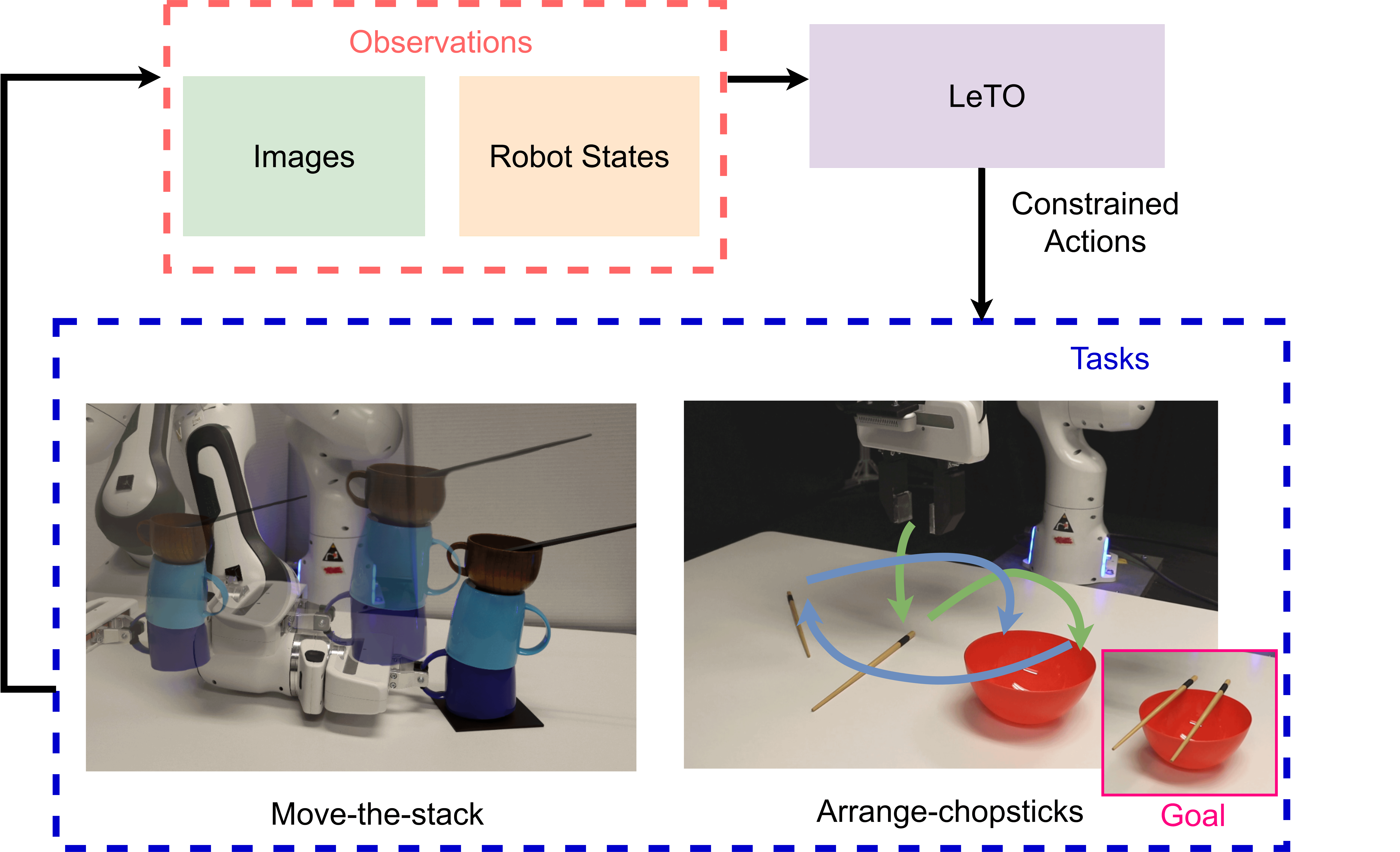}
\end{overpic}
\caption{In LeTO, we enable the model to end-to-end generate actions in a safe and constraint-controlled fashion without extra modules. To the best of our knowledge, LeTO is the first visuomotor imitation learning framework that not only utilizes differentiable optimization but also demonstrates its efficacy in real-world robotic manipulation tasks.}
\label{fig:firstPage}
\end{figure}

In this paper, we propose LeTO, learning constrained visuomotor policy with differentiable trajectory optimization. The strengths and novelties of our method are as follows.

1) \textbf{Differentiable optimization layer:} We integrate a differentiable optimization layer into the neural network and formulate it as a trajectory optimization problem. We demonstrate that the proposed differentiable optimization layer is feasible for optimization during training and capable of effective model representation. To the best of our knowledge, our work is the first visuomotor imitation learning framework that not only utilizes differentiable optimization but also demonstrates its efficacy in both simulation and real-world robotic manipulation tasks.

2) \textbf{Safe and constrained action generation:}
In our approach, we formulate the optimization layer as a trajectory optimization problem. Through this approach, the policy generates constrained trajectories, enhancing the overall safety and smoothness of robot actions. Since the model incorporates position, velocity, and acceleration constraints during end-to-end training, it ensures robot safety without compromising performance when deploying the model as a real-time policy. 
Compared to the ``black box" characteristics of neural networks, our approach can be described as a ``gray box" that combines the safety and interpretability of optimization-based trajectory generation with the powerful representational capabilities of neural networks. LeTO balances the objectives of skill learning and trajectory optimization with constraint guarantees. 

In Section \ref{sec:app}, we will introduce the specific methodologies of LeTO, followed by the presentation of experimental results from simulations and real-world scenarios in Sections \ref{sec:simu} and \ref{sec:real}, respectively. Section \ref{sec:diss} will be the discussion section and will also propose future research questions that are promising and intriguing in our opinion.

\section{Related work}
\subsection{Trajectory Optimization}

Trajectory optimization is pivotal in generating safe and smooth trajectories for robots and remains an active area of research. CHOMP \cite{zucker2013chomp} approaches trajectory optimization by gradient techniques, focusing on optimizing the trade-off between safety and smoothness. TrajOpt \cite{schulman2014motion} uses sequential convex optimization to generate smooth and collision-free trajecotries and can use naive straight-line initializations that might be in collision. The minimum-snap trajectory optimization  \cite{mellinger2011minimum} uses piecewise polynomials to represent the  trajectory and is optimized by quadratic programs. The work in \cite{zhang2020optimization} proposes a method for
reformulating nondifferentiable collision avoidance constraints
into smooth, differentiable constraints and enables real-time optimization-based trajectory
planning.

In LeTO, we innovatively combine trajectory optimization with imitation learning by differentiable optimization, to generate smooth and constraint-compliant trajectories during policy rollout. Our model is trained end-to-end. During the training process, LeTO not only learns the policy from human demonstrations but also learns the appropriate trajectory optimization parameters suitable for the policy. {Moreover, we would like to clarify that our paper does not introduce a new trajectory generation algorithm. Rather, it focuses on presenting an imitation learning algorithm. The entire discussion within our paper is centered on this aspect of imitation learning.}

\subsection{Imitation Learning}
The basic way of imitation learning explicitly maps observations to actions \cite{pomerleau1988alvinn, zhang2018deep,florence2019self,bojarski2016end,ross2011reduction,rahmatizadeh2018vision}. It can be trained using regression loss. However, these policies are not ideal for capturing multi-modal distributions \cite{florence2022implicit}.

Previous works have aimed to represent multi-modal distributions by  converting the regression into classification \cite{wu2020spatial,zeng2021transporter,avigal2022speedfolding}, using action discretization coupled with a multi-task action correction \cite{shafiullah2022behavior}, using MDNs \cite{mandlekar2022matters}, and using implicit modeling that including energy-based model \cite{florence2022implicit,jarrett2020strictly} and diffusion model \cite{chi2023diffusionpolicy}.  However,
these methods overlook critical aspects of safety for robotic systems. For robotic systems, their behaviors must meet specific constraints to ensure the safety of the system and the successful completion of tasks.

LeTO focuses on combining explainable and model-based trajectory optimization with imitation learning. In this way, LeTO can be end-to-end trained and generate action in a safe and constraint-controlled fashion without extra modules.

\subsection{Differentiable Optimization in Robot Learning}

Previous works utilize differentiable optimization to model discontinuous functions and dynamics \cite{Pfrommer2020,Bianchini2022}. Furthermore, some works focus on combining model-based methods with neural networks through differentiable optimization, such as using differentiable MPC \cite{amos2018differentiable} and koopman operator \cite{retchin2023koopman}. However, these methods are not suitable for tasks with high-dimensional observations, such as using camera images from multiple viewpoints. 

The work in \cite{xu2024letac} proposes a tactile-reactive grasping controller that combines image encoder and differentiable MPC. However, it cannot be used for more general policy learning.

For robot navigation and obstacle avoidance, various end-to-end learning frameworks are proposed that are embedded with differentiable optimization, such as the use of the control barrier function \cite{xiao2021barriernet}, gradient-based correction \cite{diehl2022differentiable}, and stack prediction, planning, and control in a differentiable way \cite{karkus2023diffstack}. However, all of these methods are not suitable for performing manipulation tasks.

DiffTOP \cite{wan2024difftop} is a method that integrates differentiable trajectory optimization as the policy representation to generate actions. In integrating imitation learning, DiffTOP focuses on capturing multi-modal distributions. However, it cannot generate safe and constrained trajectories. 
In contrast, LeTO focuses on how the integration of differentiable trajectory optimization into the imitation learning framework enables the generation of trajectories that not only perform tasks effectively but also satisfy constraints.

Another category of methods closely related to LeTO is riemannian motion policy (RMP) \cite{ratliff2018riemannian} and its extensions \cite{li2021rmp2,cheng2020rmp}. Specifically, the works in \cite{li2021rmp2,cheng2020rmp} enhance RMP's integration into end-to-end robot learning by incorporating automatic differentiation. This enables end-to-end training and inference for policies with RMP structures. Methods based on RMP are particularly focused on generating motion in spaces with high degrees of freedom and nonlinear dynamics. In contrast, LeTO offers a capable form of generating constrained motion at the task space trajectory level with differentiable optimization. This advantage makes it better suited for adaptation to visuomotor policies and demonstrates superior performance in real tasks. To the best of our knowledge, our work is the first visuomotor imitation learning framework that not only utilizes differentiable optimization but also demonstrates its efficacy in both simulation and real-world robotic manipulation tasks.

\section{Approach}\label{sec:app}
\begin{figure*}[t]
\centering
\begin{overpic}[trim=0 0 0 0,clip, width=1\textwidth]{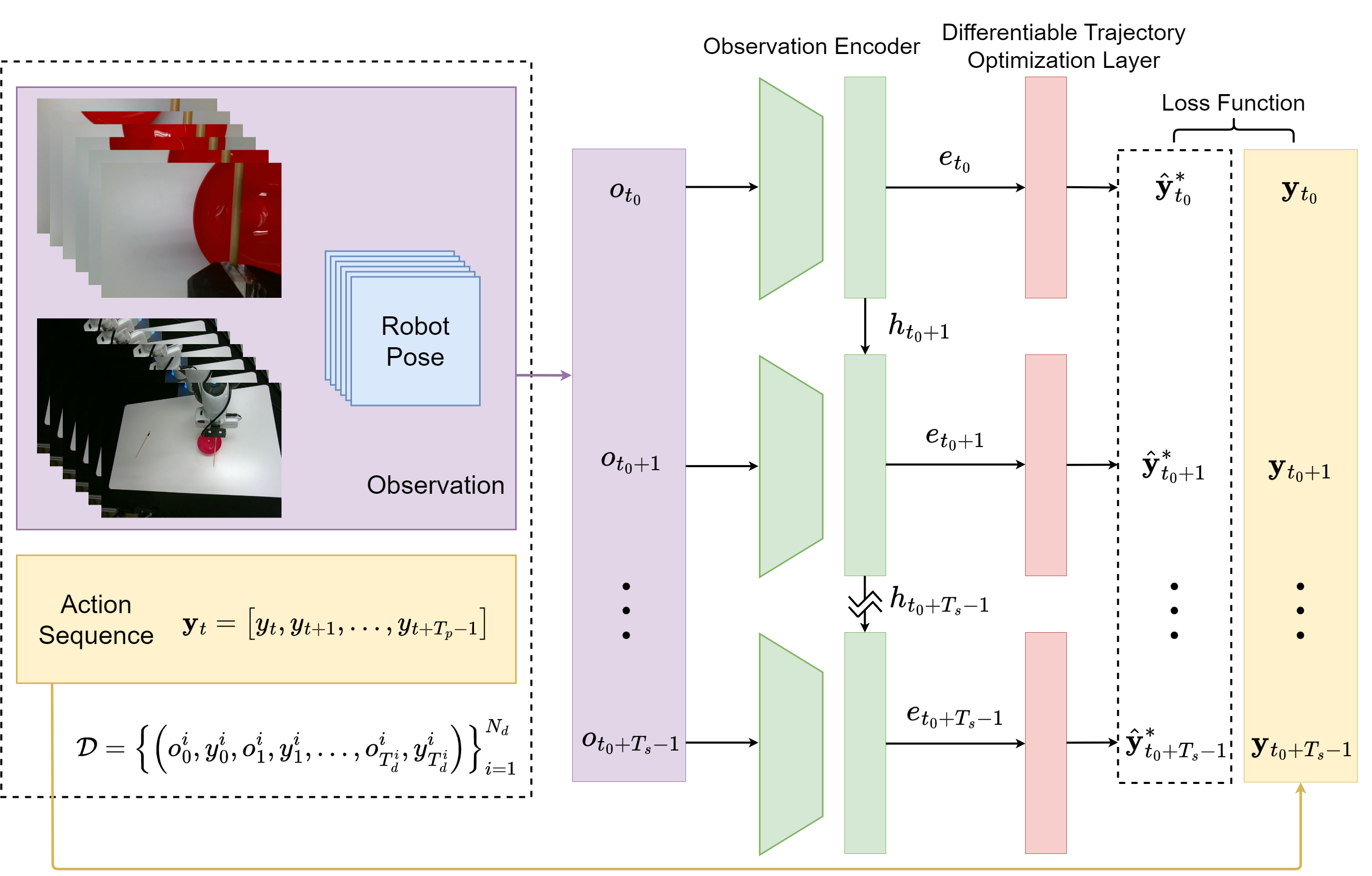}
\end{overpic}
\caption{Overview of LeTO. {We enable the model to end-to-end generate actions in a safe and constraint-controlled fashion by integrating a differentiable trajectory optimization layer. }}
\label{fig:leto_model}
\end{figure*}

In this paper, we assume access to an offline trajectories dataset of the task we want to perform. The goal is to learn a policy from this dataset offline and can successfully carry out the task by running the policy online. The overview of LeTO is shown in Fig.~\ref{fig:leto_model}. In this section, we will detail the design of the model.

\begin{figure}[t]
\centering
\begin{overpic}[trim=10 5 0 0,clip, width=0.5\textwidth]{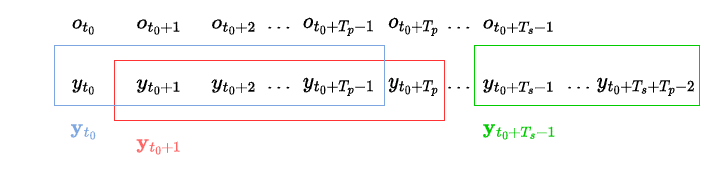}
\end{overpic}
\caption{Illustration of training data sampling.}
\label{fig:sampling}
\end{figure}

\subsection{Training Data Sampling}
In this paper, we categorize actions into two types: one is discrete actions, such as grasping and releasing; the other is continuous actions, such as the motion of the robot. For continuous actions, we consider the velocity/increment of the end-effector as the action \cite{mandlekar2022matters,mandlekar2020learning,mandlekar2020iris,shafiullah2022behavior,zhang2018deep,florence2019self}.  Denote the dataset as
$$\mathcal{D}=\left\{\left(o_0^i, y_0^i, o_1^i,y_1^i, \ldots, o_{T^i_d}^i,y_{T^i_d}^i\right)\right\}_{i=1}^{N_d},$$ where $o$ represents the observations of the robot, such as images of the cameras with different views and robot states and $y\in\mathbb{R}^{D_y}$ represents the demonstrated action aligned with the observation $o$. $N_d$ is the total number of demonstrated trajectories, $T_d$ is the total length of a trajectory, and $D_y$ is the dimension of action $y$.

Inspired by model predictive control and diffusion policy \cite{chi2023diffusionpolicy}, in LeTO, we predict an action sequence at each time step. Moreover,  we care about how to predict trajectories
that satisfy constraints rather than action sequences simply fitting the human demonstration.  Therefore, before training, we first sample many fixed-length sequences in following format from the dataset
$$
\left(o_{t_0}, \mathbf{y}_{t_0},o_{{t_0}+1}, \mathbf{y}_{{t_0}+1}, \ldots, o_{t_0+T_s-1}, \mathbf{y}_{t_0+T_s-1}\right)
,$$ where 
$\mathbf{y}_t = \left[
{y}_t^{\mathrm{T}},{y}_{t+1}^{\mathrm{T}},\ldots,{y}_{t+T_p-1}^{\mathrm{T}}
\right]^{\mathrm{T}}\in \mathbb{R}^{T_pD_y}
$. $T_s$ is the length of each data sampling and $T_p$ is the length of the predicted action sequence, as shown in Fig.~\ref{fig:sampling}. The learning objective is to minimize errors between the predicted $\hat{\mathbf{y}}^*_t$ and $\mathbf{y}_t$ (e.g. using MSE loss to assess the distance between them), as shown in Fig.~\ref{fig:leto_model}. During policy rollout, the robot will execute the first $T_a$-step actions of $\hat{\mathbf{y}}^*_t$, and then repeat the same process based on new observations. This process is typically called receding horizon planning/control.

Continuous actions need to be constrained to ensure actions' safety and smoothness, while discrete actions do not. Therefore, we define a selection matrix $S$ to pick the continuous action vector $v$ that must be constrained. 
\begin{align}
\label{eq:select}
v = Sy\in\mathbb{R}^{D_v}.
\end{align}
As mentioned earlier, we focus on a velocity-based action space, so here $v$ refers to the velocity at which the robot is being controlled {and $D_v$ is the dimension of $v$}.

\subsection{Observation Encoder}

The observation encoder maps observations, such as raw RGB images and robot states, to
a latent embedding $e$ and is trained end-to-end with LeTO. To capture the sequential correlation of the task, we use a recurrent neural network (RNN) architecture, $E_{\theta}$, where the right subscript $\theta$ represents the parameters of the network. We choose long short-term memory (LSTM) as our RNN model. At each time step, $e_t, h_{t+1}=E_\theta\left(o_t, h_t\right),$ where $h$ is the hidden state. We define $e_t \in \mathbb{R}^{T_pD_y}$. More detailed reasons of setting the dimension as $T_pD_y$ can be seen in Section~\ref{subsec:dto}.

\subsection{Differentiable Trajectory Optimization Layer}\label{subsec:dto}

The differentiable trajectory optimization (DTO) layer maps the embedding $e_t$ to a predicted action sequence $\hat{\mathbf{y}}_t = \left[
\hat{y}_t^{\mathrm{T}},\hat{y}_{t+1}^{\mathrm{T}},\ldots,\hat{y}_{t+T_p-1}^{\mathrm{T}}
\right]^{\mathrm{T}}\in \mathbb{R}^{T_pD_y}
$. In this section, we will show how to design this layer based on trajectory optimization, learn this layer during training, and integrate it with the entire neural network.  

By equation \eqref{eq:select}, the predicted velocity command sequence 

\begin{align}\hat{\mathbf{v}}_t = \mathbf{S}\hat{\mathbf{y}}_t = \left[
\hat{v}_t^{\mathrm{T}},\hat{v}_{t+1}^{\mathrm{T}},\ldots,\hat{v}_{t+T_p-1}^{\mathrm{T}}
\right]^{\mathrm{T}} \in \mathbb{R}^{T_pD_v},\label{eq:select2}
\end{align}
where $\mathbf{S} = \text{blkdiag}\left(S,\dots,S \right)
    \in \mathbb{R}^{T_pD_v \times T_pD_y}$. Then
the predicted acceleration sequence and position sequence are 

\begin{align}
\hat{\mathbf{a}}_t &= \left[\hat{a}_t^{\mathrm{T}},\hat{a}_{t+1}^{\mathrm{T}},\ldots,\hat{a}_{t+T_p-2}^{\mathrm{T}}\right]^{\mathrm{T}} \notag\\
 & = \frac{1}{\Delta t}\left[\hat{v}_{t+1}^{\mathrm{T}}-\hat{v}_{t}^{\mathrm{T}},\ldots,\hat{v}_{t+T_p-1}^{\mathrm{T}} - \hat{v}_{t+T_p-2}^{\mathrm{T}}\right]^{\mathrm{T}}\notag\\
 & =  \frac{1}{\Delta t}\left[ \begin{matrix}
-1 & 1 &  &  &  \\
 & -1 & 1 &  &  \\
 &  & \ddots & \ddots &  \\
 &  &  & -1 & 1 
\end{matrix}  \right]\hat{\mathbf{v}}_t\notag\\
 & = \mathbf{A}_{\text{diff}} \hat{\mathbf{v}}_t \in \mathbb{R}^{T_pD_v-D_v},\label{eq:acc}\\
\hat{\mathbf{p}}_t &= \left[\hat{p}_{t+1},\ldots,\hat{p}_{t+T_p-1}\right]^{\mathrm{T}}\notag\\
                    & = 
\left[ 
\begin{matrix}
p_t   \\
p_t   \\
\vdots  \\
p_t 
\end{matrix}  
\right] + 
\left[ 
\begin{matrix}
\Delta t &  &  &    \\
\Delta t& \Delta t &    \\
 \vdots& \vdots & \ddots &  \\
\Delta t & \Delta t & \ldots &  \Delta t 
\end{matrix}  
\right]\hat{\mathbf{v}}_t\notag\\
&= \mathbf{p}_{t,0} + \mathbf{A}_{\text{inte}} \hat{\mathbf{v}}_t\in \mathbb{R}^{T_pD_v-D_v},
\label{eq:posi}
\end{align}
where 
$\hat{p}_{t+i} = {p}_{t} + \hat{v}_{t}\Delta t+\dots+\hat{v}_{t+i-1}\Delta t, i = 1,\ldots,T_p-1$, and $p_t$ is the robot end-effector position at time step $t$.

The sequence $\hat{\mathbf{p}}_t, \hat{\mathbf{v}}_t, \text{and}~\hat{\mathbf{a}}_t$ together constitutes the trajectory generated by the policy. To optimize the trajectory, we must consider the relationship between $\hat{\mathbf{p}}_t, \hat{\mathbf{v}}_t, \text{and}~\hat{\mathbf{a}}_t$ and human demonstration, as well as the constraints of $\hat{\mathbf{p}}_t, \hat{\mathbf{v}}_t, \text{and}~\hat{\mathbf{a}}_t$. Based on this, the forward pass of the DTO layer is the following optimization problem:

\begin{align}
\hat{\mathbf{y}}^{\star}_t&=\underset{\hat{\mathbf{y}}_t}{\operatorname{argmin}} \frac{1}{2} \hat{\mathbf{y}}_t^{\mathrm{T}} \mathbf{Q} \hat{\mathbf{y}}_t+e_t^{\mathrm{T}} \hat{\mathbf{y}}_t +\frac{\alpha}{2} \hat{\mathbf{a}}_t^{\mathrm{T}} \hat{\mathbf{a}}_t, \label{eq:opt1}\\
\text { subject to }~&b_{\text{min}} \leq A_{\text{pos}} \hat{p}_{t+i} \leq b_{\text{max}},i = 1,\ldots,T_p-1,\notag\\
&v_{\text{min}}\leq \hat{v}_{t+j} \leq {v}_{\text{max}},j = 0,\ldots,T_p-1,\notag\\
 &a_{\text{min}} \leq \hat{a}_{t+k} \leq a_{\text{max}}, k = 0, \ldots T_p-2.\notag
\end{align}
Rewrite the optimization problem \eqref{eq:opt1} by equations \eqref{eq:select2}, \eqref{eq:acc}, and \eqref{eq:posi}, then we have
\begin{align}
\hat{\mathbf{y}}^{\star}_t&=\underset{\hat{\mathbf{y}}_t}{\operatorname{argmin}} \frac{1}{2} \hat{\mathbf{y}}_t^{\mathrm{T}} \bar{\mathbf{Q}} \hat{\mathbf{y}}_t+e_t^{\mathrm{T}} \hat{\mathbf{y}}_t,\label{eq:opt2}\\
\text { subject}&~\text{to}\notag\\
&\mathbf{b}_{\text{min}} \leq \mathbf{A}_{\text{pos}} (\mathbf{p}_{t,0} + \mathbf{A}_{\text{inte}} \mathbf{S}\hat{\mathbf{y}}_t) \leq \mathbf{b}_{\text{max}},\label{eq:posi_con}\\
&\mathbf{v}_{\text{min}}\leq \mathbf{S}\hat{\mathbf{y}}_{t} \leq \mathbf{v}_{\text{max}},\label{eq:vel_con}\\
 &\mathbf{a}_{\text{min}} \leq \mathbf{A}_{\text{diff}}\mathbf{S}\hat{\mathbf{y}}_t \leq \mathbf{a}_{\text{max}},\label{eq:acc_con}\\
 \text{where}&~\bar{\mathbf{Q}} = (\mathbf{Q}+\alpha \mathbf{S}^\mathrm{T}\mathbf{A}_{\text{diff}}^\mathrm{T}\mathbf{A}_{\text{diff}} \mathbf{S}), \notag\\
 &~\mathbf{A}_{\text{pos}} = \text{blkdiag}\left(A_{\text{pos}} ,\dots,A_{\text{pos}}  \right) \notag\\
    &~~~~~~~~\in \mathbb{R}^{(T_pD_v-D_v) \times (T_pD_v-D_v)}. \notag
\end{align}

The coefficient $\alpha$  controls the smoothness of the trajectory. It can be observed in \eqref{eq:opt1} that the larger the value of $\alpha$, the higher the cost associated with the sum of accelerations, leading the optimization problem to generate trajectories with a smaller sum of accelerations. $A_{\text{pos}}$ and $b_{\text{max,min}}$ define the convex constraints that the position must satisfy. $v_{\text{max,min}}$ and $a_{\text{max,min}}$ define the constraints for the velocity and acceleration of the trajectory. $\alpha,A_{\text{pos}},b_{\text{max,min}},v_{\text{max,min}}$ and $a_{\text{max,min}}$ are interpretable and model-based, so we do not need to learn these parameters during training. Instead, it is advisable to specify them according to the requirements of the task. In contrast, $\mathbf{Q}\in \mathbb{R}^{T_pD_y}$ is a parameter that we need to learn during training. Next, we will demonstrate how to learn $\mathbf{Q}$ and its specific meaning in our trajectory optimization problem.

Optimization problem~\eqref{eq:opt2} is a quadratic program (QP). Therefore,
the layer~\eqref{eq:opt2} can perform forward pass and
backpropagation in batch form as long as it is always feasible in the training process \cite{amos2017optnet}. The feasibility of \eqref{eq:opt2} can be guaranteed by ensuring $\mathbf{Q}$ symmetric positive definite \cite{boyd2004convex}. To achieve that, we use a Cholesky factorization
$$
\mathbf{Q}=\mathbf{L} \mathbf{L}^{\mathrm{T}} + \epsilon \mathbf{I},
$$
and directly learn $\mathbf{L}$ , where $\mathbf{L}$ is a lower triangular matrix and $\epsilon$ is a very small scalar (e.g. $1\times10^{-4}$). By observing optimization problem \eqref{eq:opt2}, Remark~\ref{rm:1} can be derived.

\begin{remark}[feasibility of the differentiable optimization layer]
\label{rm:1}
If the constraint $\mathbf{b}_{\rm{min}} \leq \mathbf{A}_{\rm{pos}} \mathbf{p}_{t,0} \leq \mathbf{b}_{\rm{max}}$ is satisfied at each time step $t$, the optimization problem is always feasible, regardless of how $\mathbf{L} $ and $e_t$ change. This means that the optimization problem remains consistently feasible, which ensures a stable training process. $\mathbf{b}_{\rm{min}} \leq \mathbf{A}_{\rm{pos}} \mathbf{p}_{t,0} \leq \mathbf{b}_{\rm{max}} $ represents that the human demonstrations should satisfy the position constraints. Additionally, human demonstrations do not need to satisfy the velocity and acceleration constraints strictly, because they have nothing to do with the feasibility of the optimization problem. Further more, for human demonstration, satisfying position constraints is much easier than satisfying velocity and acceleration constraints. 
\end{remark}

Based on the fact that $\mathbf{Q}$ is symmetric positive definite, the optimization problem \eqref{eq:opt2} can be converted to the following least squares problem:

\begin{align}
 &\hat{\mathbf{y}}^{\star}_t=\underset{\hat{\mathbf{v}}_t}{\operatorname{argmin}} \frac{1}{2}\|\bar{\mathbf{L}}^{\mathrm{T}} \hat{\mathbf{y}}_t-{\bar{e}_t}\|^2,\label{eq:opt3}\\
&\text { subject to }~\eqref{eq:posi_con}, \eqref{eq:vel_con}, \text{and}~ \eqref{eq:acc_con},\notag\\
&\text { where }~\bar{\mathbf{Q}}=\bar{\mathbf{L}} \bar{\mathbf{L}}^{\mathrm{T}},\notag \\
&~~~~~~~~~~{e_t}=-\bar{\mathbf{L}} {\bar{e}_t}.\notag
\end{align}
Based on Cholesky factorization, $\bar{\mathbf{L}}$ is a lower triangular matrix  with real and positive diagonal entries. By observing optimization problem \eqref{eq:opt3}, Remark~\ref{rm:2}  can be derived.

\begin{remark}[representational power]
\label{rm:2}
DTO layer actually optimizes a linear transformation under motion constraints. Since $\bar{\mathbf{L}}$ is a lower triangular matrix and it is fully ranked, the linear transformation here is 
$$\hat{\mathbf{y}}^{\star}_t = -({\bar{\mathbf{L}}^{\mathrm{T}}})^{-1}\bar{\mathbf{L}}^{-1}e_t.$$ 
Since the observation encoder is just a normal LSTM  architecture, it is already has the power of representing the policy. Therefore, the whole model including the DTO layer has the power to fit the human demonstration.
\end{remark}

\subsection{Policy Deployment}
When the policy is deployed on the robot, the RNN is unrolled one-step at a time, 
$e_t, h_{t+1}=E_\theta\left(o_t, h_t\right)$. The
hidden state $h$ is refreshed every $T_s$ steps. In addition, the training is done on discrete trajectory clips while the robot trajectory is continuous in reality. Therefore, the forward pass of the DTO layer need to add a new constraint during test time to fully constrain the motion generation. 

\begin{align*}
\hat{\mathbf{y}}^{\star}_t=\underset{\hat{\mathbf{y}}_t}{\operatorname{argmin}} &\frac{1}{2} \hat{\mathbf{y}}_t^{\mathrm{T}} \bar{Q} \hat{\mathbf{y}}_t+e_t^{\mathrm{T}} \hat{\mathbf{y}}_t,\\
\quad \text { subject to }&~\eqref{eq:posi_con}, \eqref{eq:vel_con}, \text{and}~\eqref{eq:acc_con},\\
 a_{\text{min}}\Delta t &\leq \hat{v}_{t} - \hat{v}_{t-1} \leq  a_{\text{max}}\Delta t 
\end{align*}
where $\hat{v}_{t-1}$ represents the last executed action for the robot at time step $t$ (executed at time step $t-1$). Adding this new constraint is to transfer the model trained on a discrete set of trajectories to continuous online trajectory optimization.

\begin{table*}[t]
\caption{{Experimental task setups and configurations. \textbf{Initial States}: Is the initial configuration (position and orientation) of the manipulated object randomized? \textbf{Objective}: Is the target object in the manipulation task randomized? For example, in the Square task, the stick to insert the nut remains stationary. \textbf{PH and MH}: The number of demonstrations in the dataset. PH represents proficient-human demonstrations, while MH stands for multi-human demonstrations. Our real-world task experiments were conducted exclusively on the PH dataset, consistent with the setup described in \cite{chi2023diffusionpolicy}.  \textbf{Act. Dim.}: Action dimensions. \textbf{HiPrec}: Is high-precision manipulation required? \textbf{Multi-Step}: Does the task involve multiple sub-tasks? \textbf{Num. of Cam.}: The number of cameras used.}}
\centering
{
\begin{tabular}{|c|c|c|c|c|c|c|c|c|}
\hline
\textbf{Task}               & \textbf{Initial States} & \textbf{Objective} & \textbf{PH} & \textbf{MH} & \textbf{Act. Dim.} & \textbf{HiPrec} & \textbf{Multi-Steps} & \textbf{Num. of Cam.} \\ \hline
\textbf{Can}                & Random                & Fixed              & 200         & 300         & 7                  & No               & No                    & 2                  \\
\textbf{Square}             & Random                & Fixed              & 200         & 300         & 7                  & Yes              & No                    & 2                  \\
\textbf{Move-the-stack}     & Random                & Random             & 100         & -           & 3                  & Yes              & No                    & 1                  \\
\textbf{Arrange-chopsticks} & Random                & Fixed              & 120         & -           & 4                  & No               & Yes                   & 2                  \\ \hline
\end{tabular}}
\label{tb:task_config}
\end{table*}

\begin{table*}[t]
\caption{{Training configuration of LeTO in simulation benchmarks. To improve learning efficiency, actions processed through LeTO are first normalized to a range between -1 and 1 \cite{mandlekar2022matters}. Consequently, the velocity and acceleration constraints applied here pertain to the normalized actions.}}
\centering
{
\begin{tabular}{|c|c|c|c|c|}
\hline
\textbf{Velocity Constraints} [${v}_{\text{min}},{v}_{\text{max}}$] & \textbf{Acceleration Constraints} [${a}_{\text{min}},{a}_{\text{max}}$]& \textbf{Smoothing Weight} $\alpha$ & \textbf{Sampling Length} $T_s$ & {\textbf{Predicted Length} $T_p$} \\ \hline
{[}-1,1{] (normalized values)}           & {[}-0.1,0.1{] (normalized values)}           & 1                & 12              & {6}   \\\hline            
\end{tabular}}
\label{tb:train_config}
\end{table*}

\begin{table*}[ht]
\centering
\caption{Summary of success rates. We use the imitation learning benchmark, RoboMimic (Visual Policy) \cite{mandlekar2022matters}. We present the average of the maximum  success rate (\%) across 3 training seeds and 25 different initial environmental conditions (totaling 75 conditions). Results for IBC were sourced from \cite{chi2023diffusionpolicy} to enable a comparison with conventional methods.  The black bold font indicates the highest success rate among all tasks. Notably, for the square task, LeTO achieves a comparable success rate to diffusion policy (bold in green) and surpasses diffusion policy with constraints clipping.}
\begin{tabular}{|c|c|c|c|c|}
\hline &  \multicolumn{2}{|c|}{ Can } & \multicolumn{2}{|c|}{ Square } \\
\hline &  $\mathrm{ph}$ & $\mathrm{mh}$ & $\mathrm{ph}$ & $\mathrm{mh}$ \\
\hline 
IBC \cite{florence2022implicit}  & 8  & 0 & 3 & 0    \\
LSTM-GMM \cite{mandlekar2022matters}   & $94.6\pm2.3$ & $94.6\pm2.3$ & $78.7\pm6.1$ &  $77.3\pm8.3$   \\
LSTM-GMM clipping   &$90.7\pm4.6$ & $89.3\pm2.3$ &$73.3\pm8.3$  & $70.7\pm2.3$  \\
DiffusionPolicy \cite{chi2023diffusionpolicy} & $\textbf{100.0}\pm\textbf{0.0}$ & $\textbf{100.0}\pm \textbf{0.0} $ & $\textbf{98.7}\pm \textbf{2.3} $ &  $\textbf{90.7}\pm\textbf{4.6}$ \\
DiffusionPolicy clipping  & $98.7\pm2.3$ & $93.3\pm4.6$   &  $92.0\pm4.0$ & $82.7\pm2.3$  \\
LeTO  & $\textbf{100.0}\pm\textbf{ 0.0}$ & $\textbf{100.0}\pm \textbf{0.0}$ & $\textbf{\green{94.7}}\green{\pm}\textbf{\green{4.6}}$ & $\textbf{\green{88.0}}\green{\pm}\textbf{\green{4.0}}$ \\
\hline
\end{tabular}
\label{tb:succ_rate_sim}
\end{table*}

\begin{figure}[t]
\centering
\begin{overpic}[trim=440 0 0 0,clip, width=0.48\textwidth]{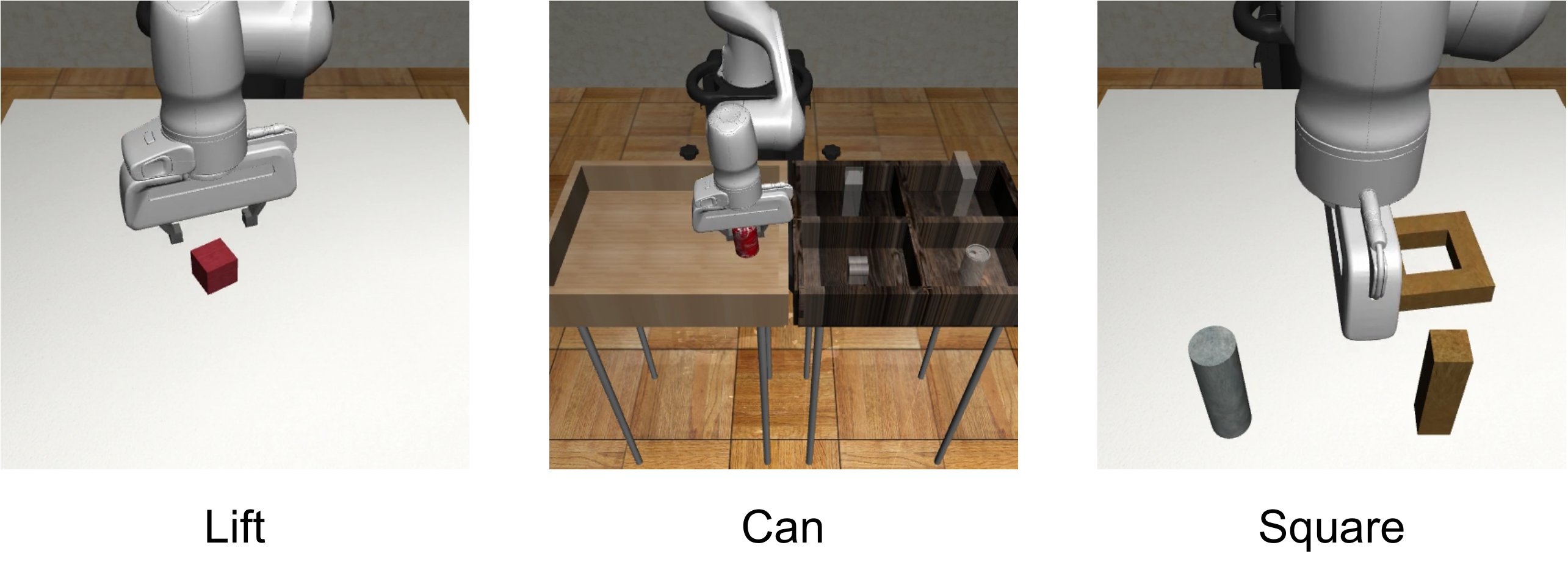}
\end{overpic}
\caption{Simulation benchmarks: pick and place the can (can task) and grasping and assembling the square nut (square task).}
\label{fig:sim_env}
\end{figure}

\section{Simulation Evaluation}\label{sec:simu}

\subsection{Experimental Setup}

We employed two simulation tasks used in \cite{mandlekar2022matters} for our simulation evaluation: pick and place the can (can task) and grasping and assembling the square nut (square task), as shown in Fig.~\ref{fig:sim_env}. More details of the task configurations can be seen in Table~\ref{tb:task_config}. The results demonstrate that within these two simulation tasks, LeTO not only achieves success
rates comparable to diffusion policy \cite{chi2023diffusionpolicy}, but also significantly surpasses IBC \cite{florence2022implicit} and LSTM+GMM \cite{mandlekar2022matters}. Furthermore, owing to the presence of constraints and differentiable trajectory optimization, our method exhibits superior performance in terms of generated trajectory quality compared to other methods, which is very important for robot systems.

{For training parameters of diffusion policy and LSTM-GMM, we use the default configurations provided in their released code. We also do the same for our realworld evaluation. We have released our code, training data, and checkpoints for reproducing our results.}

The can and square tasks are from the RoboMimic benchmark \cite{mandlekar2022matters}. Apart from the most basic lifting task, we opted not to include the tool hanging and transport tasks. As highlighted in \cite{amos2017optnet}, solving optimization problems precisely, as we are doing here, exhibits cubic complexity concerning the number of variables and/or constraints. Consequently, our method's training speed for tasks like tool hanging and transport, which involve larger datasets or high-dimensional output, is sluggish, and under limited GPU computational resources, obtaining results within a short time is unfeasible. Nevertheless, as we have already demonstrated the capability of our approach to achieve high successful rates and generate high-quality trajectories through the outcomes of the two tasks mentioned above, we decided against further training on the tool hanging and transport tasks.

We train policies on 3 seeds (42, 142, 242) with batch size 64, and test in 25 different initial environments for each seed (totaling 75 conditions). For each task, we train policies on two different types of datasets: Proficient-Human (ph), and Multi-Human (mh) datasets \cite{mandlekar2022matters}. We report the success rate (see Table~\ref{tb:succ_rate_sim}) of all these experiments. We also report metrics to evaluate the quality of the generated trajectories (see Tables~\ref{tb:can_ph}, \ref{tb:can_mh}, \ref{tb:square_ph}, and~\ref{tb:square_mh}).

\begin{table*}[ht]
\caption{Trajectory metrics ($10^{-2}$) of the can-ph task. The results correspond to the results in Table~\ref{tb:succ_rate_sim}. {These values represent normalized accelerations. To convert the results back to their raw form, use a scalar of 20 $\text{m}/\text{s}^2$ for linear acceleration and 200 $\text{rad}/\text{s}^2$ for rotational acceleration. For example, the maximum linear acceleration for LSTM-GMM is $10^{-2} \times 77.49 \times 20~\text{m}/\text{s}^2 = 15.50~\text{m}/\text{s}^2 $ (which exceeds the gravitational acceleration), whereas for LeTO it is only $10^{-2} \times 10 \times 20~\text{m}/\text{s}^2 = 2~\text{m}/\text{s}^2$.}}
\centering
\begin{tabular}{|c|c|c|c|c|c|c|c|}
\hline 
& {avg-mean-lin} & {avg-max-lin} & {avg-std-lin} & {avg-mean-rot} & {avg-max-rot} & {avg-std-rot} &num-of-opt \\
\hline 
DiffusionPolicy \cite{chi2023diffusionpolicy}& $8.23\pm 0.70$ &$69.78\pm 2.67$ & $8.39\pm 0.58$  &$\textbf{1.03}\pm\textbf{ 0.12}$ & $12.73\pm 0.78$ & $\textbf{1.29}\pm \textbf{0.12}$&2\\
LSTM-GMM \cite{mandlekar2022matters}& $9.78\pm 3.54$ &$77.49\pm 15.16$ & $10.61\pm 3.48$  &$\textbf{1.04}\pm \textbf{0.27}$ & $24.58\pm 9.73$ & $2.71\pm 0.26$&1\\
DiffusionPolicy clipping & $7.73\pm 0.10$ &$\textbf{10.00}\pm \textbf{0.00}$& $\textbf{3.21}\pm\textbf{ 0.06}$  &$3.61\pm 0.47$ & $\textbf{9.94}\pm \textbf{0.07}$ & $2.45\pm 0.08$&3\\
LSTM-GMM clipping & $5.61\pm 1.03$ &$\textbf{10.00}\pm \textbf{0.00}$ & $3.68\pm 0.10$  &$1.16\pm 0.13$ & $\textbf{9.59}\pm \textbf{0.14}$ & $\textbf{1.37}\pm \textbf{0.05}$&3\\
LeTO & $\textbf{4.36}\pm \textbf{0.10}$ &$\textbf{10.00}\pm \textbf{0.00}$ & $\textbf{3.47}\pm \textbf{0.03}$  &$1.50\pm 0.21$ & $\textbf{9.78}\pm \textbf{0.37}$ & $1.73\pm 0.30$&\textbf{4}\\
\hline
\end{tabular}
\label{tb:can_ph}
\end{table*}

\begin{table*}[ht]
\caption{Trajectory metrics ($10^{-2}$) of the can-mh task. The results correspond to the results in Table~\ref{tb:succ_rate_sim}. {These values represent normalized accelerations. To convert the results back to their raw form, use a scalar of 20 $\text{m}/\text{s}^2$ for linear acceleration and 200 $\text{rad}/\text{s}^2$ for rotational acceleration.}}
\centering
\begin{tabular}{|c|c|c|c|c|c|c|c|}
\hline 
& {avg-mean-lin} & {avg-max-lin} & {avg-std-lin} & {avg-mean-rot} & {avg-max-rot} & {avg-std-rot} &num-of-opt \\
\hline 
DiffusionPolicy \cite{chi2023diffusionpolicy} &  $5.97\pm 0.93$ &$64.85\pm 9.04$ & $7.32\pm 0.85$  &$\textbf{0.60}\pm \textbf{0.06}$ & $10.95\pm 1.16$ & $\textbf{0.96}\pm \textbf{0.09}$ &2\\
LSTM-GMM \cite{mandlekar2022matters}&  $6.02\pm 0.78$ &$72.16\pm 1.94$ & $7.41\pm 0.36$  &$0.75\pm 0.08$ & $21.48\pm 9.64$ & $1.57\pm 0.50$&0\\
DiffusionPolicy clipping &  $7.35\pm 0.14$ &$\textbf{10.00}\pm \textbf{0.00}$ & $3.30\pm 0.03$  &$2.74\pm 0.19$ & $9.80\pm 0.07$ & $2.26\pm 0.12$&1\\
LSTM-GMM clipping &  $4.74\pm 0.46$ &$\textbf{10.00}\pm \textbf{0.00}$ & $3.45\pm 0.10$  &$0.79\pm 0.10$ & $9.16\pm 0.87$ & $1.19\pm 0.43$&1\\
LeTO &  $\textbf{2.66}\pm \textbf{0.68}$ &$\textbf{10.00}\pm \textbf{0.00}$ & $\textbf{2.60}\pm \textbf{0.31}$  &${0.74}\pm {0.22}$ & $\textbf{8.28}\pm \textbf{1.04}$ & $\textbf{0.98}\pm \textbf{0.24}$&\textbf{5}\\
\hline
\end{tabular}
\label{tb:can_mh}
\end{table*}

\begin{table*}[ht]
\caption{Trajectory metrics ($10^{-2}$) of the square-ph task. The results correspond to the results in Table~\ref{tb:succ_rate_sim}. {These values represent normalized accelerations. To convert the results back to their raw form, use a scalar of 20 $\text{m}/\text{s}^2$ for linear acceleration and 200 $\text{rad}/\text{s}^2$ for rotational acceleration.}}
\centering
\begin{tabular}{|c|c|c|c|c|c|c|c|}
\hline 
& {avg-mean-lin} & {avg-max-lin} & {avg-std-lin} & {avg-mean-rot} & {avg-max-rot} & {avg-std-rot} &num-of-opt \\
\hline 
DiffusionPolicy \cite{chi2023diffusionpolicy} & $8.17\pm 0.78$ &$70.66\pm 4.49$ & $8.65\pm 0.40$  &$\textbf{0.95}\pm \textbf{0.11}$ & $13.22\pm 0.80$ & $\textbf{1.32}\pm \textbf{0.11}$&2\\
LSTM-GMM \cite{mandlekar2022matters} & $8.19\pm 2.49$ &$68.75\pm 6.52$ & $8.76\pm 1.82$  &$\textbf{0.91}\pm \textbf{0.08}$ & $13.81\pm 3.51$ & $\textbf{1.30}\pm \textbf{0.24}$&2\\
DiffusionPolicy clipping & $6.90\pm 0.75$ &$\textbf{10.00}\pm \textbf{0.00}$ & $\textbf{3.36}\pm \textbf{0.06}$  &$2.09\pm 0.88$ & $\textbf{9.09}\pm \textbf{0.82}$ & $1.75\pm 0.59$&3\\
LSTM-GMM clipping & $\textbf{5.11}\pm \textbf{1.46}$ & $\textbf{10.00}\pm \textbf{0.00}$ & $\textbf{3.36}\pm \textbf{0.20}$ & $\textbf{0.98}\pm \textbf{0.22}$& $\textbf{9.04}\pm \textbf{0.54}$ &$\textbf{1.27}\pm \textbf{0.29}$ &\textbf{6}\\
LeTO & $\textbf{5.08}\pm \textbf{0.33}$ &$\textbf{10.00}\pm \textbf{0.00}$ & $\textbf{3.35}\pm \textbf{0.03}$  &$1.28\pm 0.17$ & $\textbf{9.24}\pm \textbf{0.15}$ & $\textbf{1.37}\pm \textbf{0.08}$&5\\
\hline
\end{tabular}
\label{tb:square_ph}
\end{table*}

\begin{table*}[t]
\caption{Trajectory metrics ($10^{-2}$) of the square-mh task. The results correspond to the results in Table~\ref{tb:succ_rate_sim}. {These values represent normalized accelerations. To convert the results back to their raw form, use a scalar of 20 $\text{m}/\text{s}^2$ for linear acceleration and 200 $\text{rad}/\text{s}^2$ for rotational acceleration.}}
\centering
\begin{tabular}{|c|c|c|c|c|c|c|c|}
\hline 
& {avg-mean-lin} & {avg-max-lin} & {avg-std-lin} & {avg-mean-rot} & {avg-max-rot} & {avg-std-rot} &num-of-opt \\
\hline 
DiffusionPolicy \cite{chi2023diffusionpolicy}& $5.42\pm 0.30$ &$60.97\pm 4.38$ & $6.61\pm 0.44$  &$\textbf{0.57}\pm \textbf{0.08}$ & $13.41\pm 3.45$ & $1.11\pm 0.26$&1\\
LSTM-GMM \cite{mandlekar2022matters} & $5.37\pm 0.66$ &$54.43\pm 6.99$ & $6.28\pm 1.00$  &$0.64\pm 0.08$ & $13.31\pm 2.66$ & $1.04\pm 0.22$&0\\
DiffusionPolicy clipping  & $7.04\pm 0.42$ &$\textbf{10.00}\pm \textbf{0.00}$ & $3.35\pm 0.06$  &$2.41\pm 0.50$ & $9.46\pm 0.34$ & $2.04\pm 0.36$&1\\
LSTM-GMM clipping & $3.81\pm 0.44$ &$\textbf{10.00}\pm \textbf{0.00}$ & $3.07\pm 0.13$  &$\textbf{0.63}\pm \textbf{0.04}$ & $\textbf{7.80}\pm \textbf{0.23}$ & $\textbf{0.85}\pm \textbf{0.04}$&\textbf{4}\\
LeTO & $\textbf{3.08}\pm \textbf{0.04}$ &$\textbf{10.00}\pm \textbf{0.00}$ & $\textbf{2.70}\pm \textbf{0.01}$  &$0.91\pm 0.06$ & $\textbf{8.18}\pm \textbf{0.07}$ & $1.09\pm 0.07$&\textbf{4}\\
\hline
\end{tabular}
\label{tb:square_mh}
\end{table*}

\begin{table}[t]
\caption{{The results of LeTO on square-ph with varying acceleration constraints. To enhance readability, we have highlighted the optimal values in each metric in bold red and the second-optimal values in bold green. The acceleration results employ the same normalization as outlined in Tables~\ref{tb:can_ph}, \ref{tb:can_mh}, \ref{tb:square_ph}, and~\ref{tb:square_mh}}}
\centering
{
\begin{tabular}{|c|c|c|c|c|c|}
\hline
 & \textbf{0.05} & \textbf{0.1} & \textbf{0.2} & \textbf{0.5} & \textbf{1} \\ \hline
succ-rate                      & 0.88          & \textbf{\red{0.96}}         & \textbf{\red{0.96}}         & \textbf{\green{0.92}}         & 0.88       \\
\hline
avg-mean-lin                      & \textbf{\red{0.033}}         & \textbf{\green{0.049}}        & 0.061        & 0.071        & 0.067      \\
avg-max-lin                       & \textbf{\red{0.05}}          & \textbf{\green{0.10}}         & 0.20         & 0.46         & 0.58       \\
avg-std-lin                       & \textbf{\red{0.017}}         & \textbf{\green{0.033}}        & 0.051        & 0.071        & 0.072      \\
\hline
avg-mean-rot                      & \textbf{\red{0.012}}         & \textbf{\green{0.013}}        & \textbf{\red{0.012}}        & \textbf{\green{0.013}}        & \textbf{\green{0.013}}      \\
avg-max-rot                       &\textbf{\red{ 0.050}}         & \textbf{\green{0.090}}        & 0.121        & 0.145        & 0.134      \\
avg-std-rot                       & \textbf{\red{0.011}}         & \textbf{\green{0.014}}        & \textbf{\green{0.014}}        & 0.015        & 0.015      \\ \hline
\end{tabular}}
\label{tb:cons_ablation}
\end{table}

 {For the CNN+LSTM backbone of LeTO, we employ the hyperparameters selected in \cite{mandlekar2022matters}. The other hyperparameters for training LeTO are detailed  in Table~\ref{tb:train_config}. To improve learning efficiency, actions processed through LeTO are first normalized to a range between -1 and 1 \cite{mandlekar2022matters}. Consequently, the velocity and acceleration constraints applied here pertain to the normalized actions.}  {See the codebase\footnote{https://github.com/ARISE-Initiative/robosuite} for more information of action unnormalization.}  

{Setting velocity constraints within the range of [-1, 1] ensure that the trajectories generated by LeTO during both training and inference do not exceed the maximum velocity in the human demonstration data. This is a logical approach since we can assume that the maximum velocity in a high-quality human demonstration dataset is reasonable and does not lead to dangerous or unstable behavior in the robot.}

{The acceleration constraint of [-0.1, 0.1] effectively imposes additional smoothness on the trajectories. As acceleration is not directly controlled, the trajectories in the human demonstration may not be sufficiently smooth. In LeTO, imposing hard acceleration constraints optimizes the generation of higher-quality, smoother trajectories while ensuring manipulation performance.} 

{For the simulation tests, we do not apply any position constraints, as our aim is to solely compare the success rates and the smoothness of the trajectories generated by different methods.}

For the assessment of trajectory quality, we calculate the maximum acceleration, average acceleration, and acceleration standard deviation for each dimension (x,y,z, roll, pitch, and yaw) and for each environment condition (25 environments for each seed). Then we calculate the average among dimensions and environments. For example
$$\text{avg-max-lin} = \frac{1}{25}\sum_{i =1}^{25}\frac{a_{x,max}^i+a_{y,max}^i+a_{z,max}^i}{3},$$
where $i$ represents the $i$-th environment and $a$ represents acceleration. ``lin" here represents linear motion and ``rot" represents rotational motion. The remaining matrices can be computed by analogy. Additionally, it is worth noting that the acceleration values here are obtained by computing the difference of the normalized velocity. It is directly proportional to the actual acceleration, and this proportionality constant is a fixed value for all experiments. We mark the values within the range of 100\% to 110\% of the minimum value in each metric as the optimal indicators (bolded), and denote the number of optimal indicators for each method as ``num-of-opt".

In order to investigate the impact of direct action clipping on success rate and trajectory smoothness for baselines, we implement a clipping method. It involves evaluating the difference between the generated action and the previously executed action to determine whether it falls within the range of -0.1 to 0.1, which are the same constraints that we implement for LeTO. If it exceeds this range, we apply clipping to bring it within this range. This approach restricts acceleration.

\subsection{Results and Analysis}

As shown in Tables~\ref{tb:can_ph}, \ref{tb:can_mh}, \ref{tb:square_ph}, and~\ref{tb:square_mh}, it can be observed that LeTO significantly enhances trajectory smoothness by trajectory optimization. {For example, if we unnormalize the values to true values for comparison, in Table~\ref{tb:can_ph}, the maximum linear acceleration for LSTM-GMM is $10^{-2} \times 77.49 \times 20~\text{m}/\text{s}^2 = 15.50~\text{m}/\text{s}^2 $ (which exceeds the gravitational acceleration), while for LeTO it is only 2 $\text{m}/\text{s}^2$. The average acceleration of LSTM-GMM is also about twice that of LeTO. These metrics indicate that the trajectories generated by LSTM-GMM are significantly jerkier than those produced by LeTO. The same analysis applies to diffusion policy, and we can obtain similar results.}

Moreover, LeTO achieves a success rate comparable to that of diffusion policy and surpasses LSTM-GMM and IBC in these tasks. Interestingly, when clipping is applied to constrain the actions generated by the diffusion policy and LSTM-GMM, although this method sometimes increases the smoothness of the trajectory, it results in a reduction in their task success rates.

An intuitive explanation for this phenomenon is that both methods do not consider constraints during their training process; they merely fitted human demonstrations. When constraints are imposed on the actions generated by the trained models during deployment, it introduces additional factors that inevitably affect their performance. In contrast, our proposed LeTO, due to its ability to achieve end-to-end trajectory optimization while fitting demonstration data, balances the training objectives of satisfying constraints, smoothing the trajectories, and minimizing errors with demonstrations.

{While all the policies we benchmark might successfully complete these tasks in simulation, constraint guarantees are crucial metrics for robotic systems. People aim for robotic systems to operate safely and reliably over the long term, not just to complete a task once. Violating constraints and producing jerky trajectories could lead to unsafe situations because neural networks are inherently uninterpretable black boxes. Moreover, if a robot consistently executes jerky trajectories to complete tasks, it places a greater burden on the hardware. Over time, these factors undermine the robustness and stability of the entire system. }

{Historically, trajectory optimization for robots has focused on meeting constraints to enhance system robustness. In our paper, LeTO can be seen as a "gray box" that offers a certain level of interpretability and can perform end-to-end trajectory optimization with hard constraints. This represents our attempt to connect trajectory optimization with robot learning. Moreover, we demonstrated that simply constraining the policy through clipping results in a decrease in performance. Therefore, we believe it is meaningful to explore how to integrate constraints into the policy to balance the objectives of skill learning and trajectory optimization with constraint guarantees. LeTO provides a practical example of how to achieve this integration.}

{To investigate the impact of acceleration constraints on the results of LeTO, we have conducted an ablation study on acceleration constraints (see Table~\ref{tb:cons_ablation} for details). Through this ablation study, we can the following observations.}

{1. Different acceleration constraints affect the success rate of the policy. Both overly restrictive and overly lenient constraints can reduce the success rate, resulting in a trend where the success rate initially increases with the constraint values before decreasing.}

{2. Table~\ref{tb:cons_ablation} demonstrates that LeTO, when implementing tight acceleration constraints, yields trajectories with notably superior smoothness based on differentiable optimization. The acceleration constraints determines the smoothness of the resulting trajectories. Tighter acceleration constraints result in smoother trajectories. As the maximum acceleration in the constraints approaches the maximum acceleration in the dataset, The trajectories generated by LeTO progressively approximate the original trajectories in the dataset.}

{3. Specifically, Tables ~\ref{tb:square_ph} and \ref{tb:cons_ablation} show that both the unconstrained LSTM-GMM and diffusion policy, and the loosely constrained LeTO ultimately generate a maximum linear acceleration of around 0.6 and a maximum rotational acceleration of about 0.14. Under relaxed or no constraints, the trajectories generated by the learned policies resemble those in the dataset across various metrics. This indicates that the dataset trajectories' linear and rotational accelerations are approximately around 0.6 and 0.14, respectively (all values are normalized). As the constraints in LeTO transition from 1 to 0.05, the maximum linear acceleration of the trajectories it generates also transitions from 0.58 to 0.05, and the rotational acceleration transitions from 0.134 to 0.05.}

{4. This ablation study highlights LeTO's superiority in terms of interpretability and controllability. When adjusting constraints, the metrics of the trajectories it generates remain consistent with the design of the differentiable trajectory optimization layer. Due to the black-box nature of neural networks, they lack the capability to ``control" the final outputs, being limited only to fitting datasets. LeTO's adjustable influence on trajectories stems from the theoretical fulfillment of constraints. This ensures that regardless of the input observations, the trajectories output by LeTO will always satisfy the given constraints by trajectory optimization. Such theoretical assurances enhance the interpretability of the entire model and are crucial for long-term safety and robustness.}

\section{Realworld Evaluation}\label{sec:real}
\begin{figure*}[t]
\centering
\begin{overpic}[trim=0 0 0 0,clip, width=0.98\textwidth]{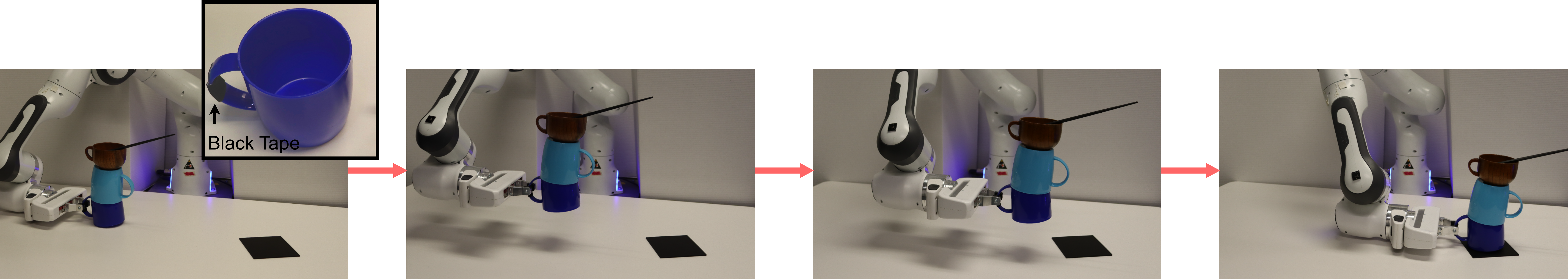}
\end{overpic}
\caption{Move-the-stack task. The robot grasps a set of stacked objects, smoothly transport them, and ultimately place them onto a black board, ensuring none of the stacked objects fall off. The black tape is for marking a consistent grasping position.}
\label{fig:stack_overview}
\end{figure*}

\begin{figure*}[t]
\centering
\begin{overpic}[trim=0 0 0 0,clip, width=0.99\textwidth]{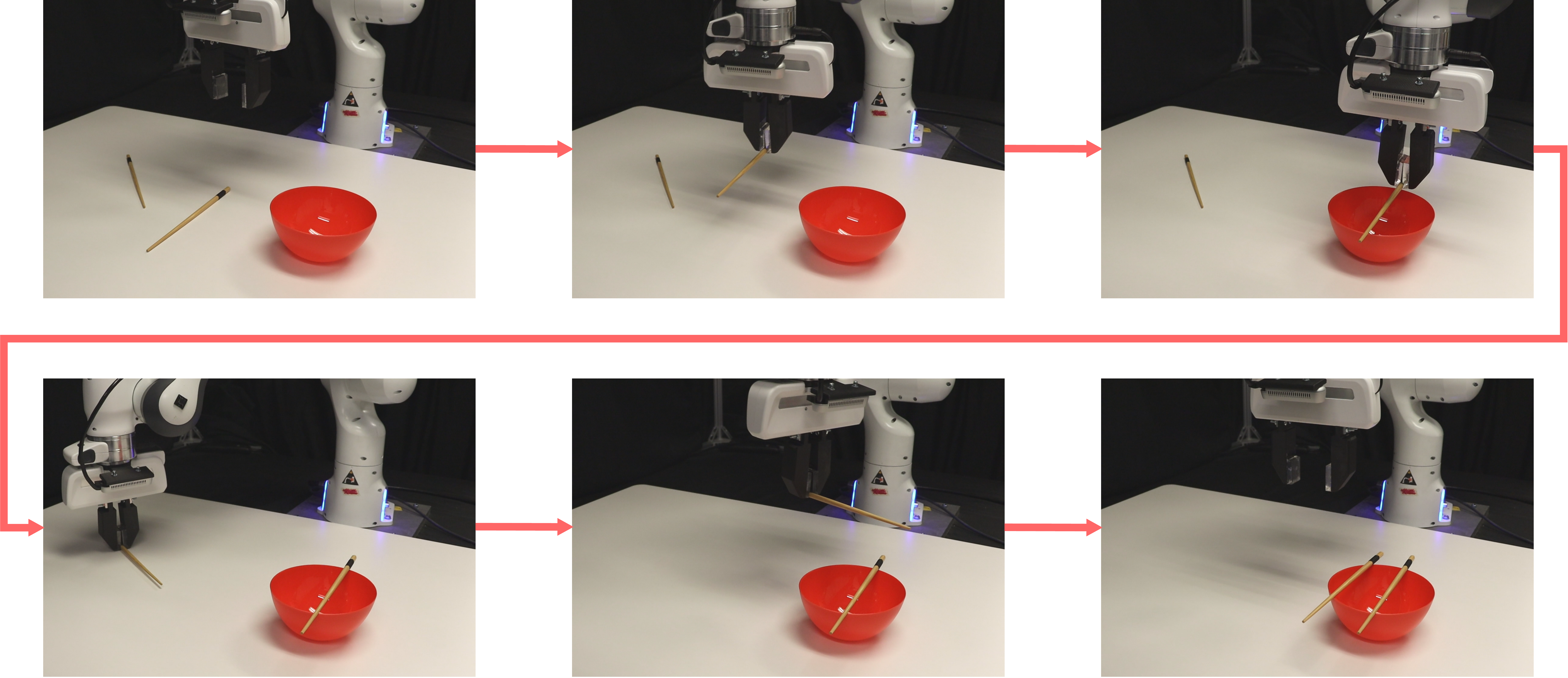}
\end{overpic}
\caption{Arrange-chopsticks task. {The robot picks up randomly placed two chopsticks and arrange them neatly on a bowl.}}
\label{fig:chop_overview}
\end{figure*}

For real robot experiments, we tackled constraints-critical tasks that demand smooth and safe trajectories, which are Move-the-stack (Fig.~\ref{fig:stack_overview}) and Arrange-chopsticks (Fig.~\ref{fig:chop_overview}). {We use SpaceMouse\footnote{https://3dconnexion.com/dk/product/spacemouse-compact/} teleoperating the end-effector of the robot manipulator to collect human demonstration data. SpaceMouse is widely used for data collection in imitation learning-related research \cite{chi2023diffusionpolicy,mandlekar2022matters,zhu2022viola,liu2022robot}. By manipulating the SpaceMouse, we can control the six degrees of freedom in the robot's end-effector velocity, as well as the gripper's grasping and releasing actions, thereby enabling the collection of data for various manipulation tasks. During the data collection process, the robot state information and images from the cameras are recorded simultaneously. More details of the task configurations can be seen in Table~\ref{tb:task_config}.}  More details of experiments and results can be seen in the attached video.

\subsection{Move-the-stack}

\begin{figure}[t]
\centering
\begin{overpic}[trim=0 0 0 0,clip, width=0.49\textwidth]{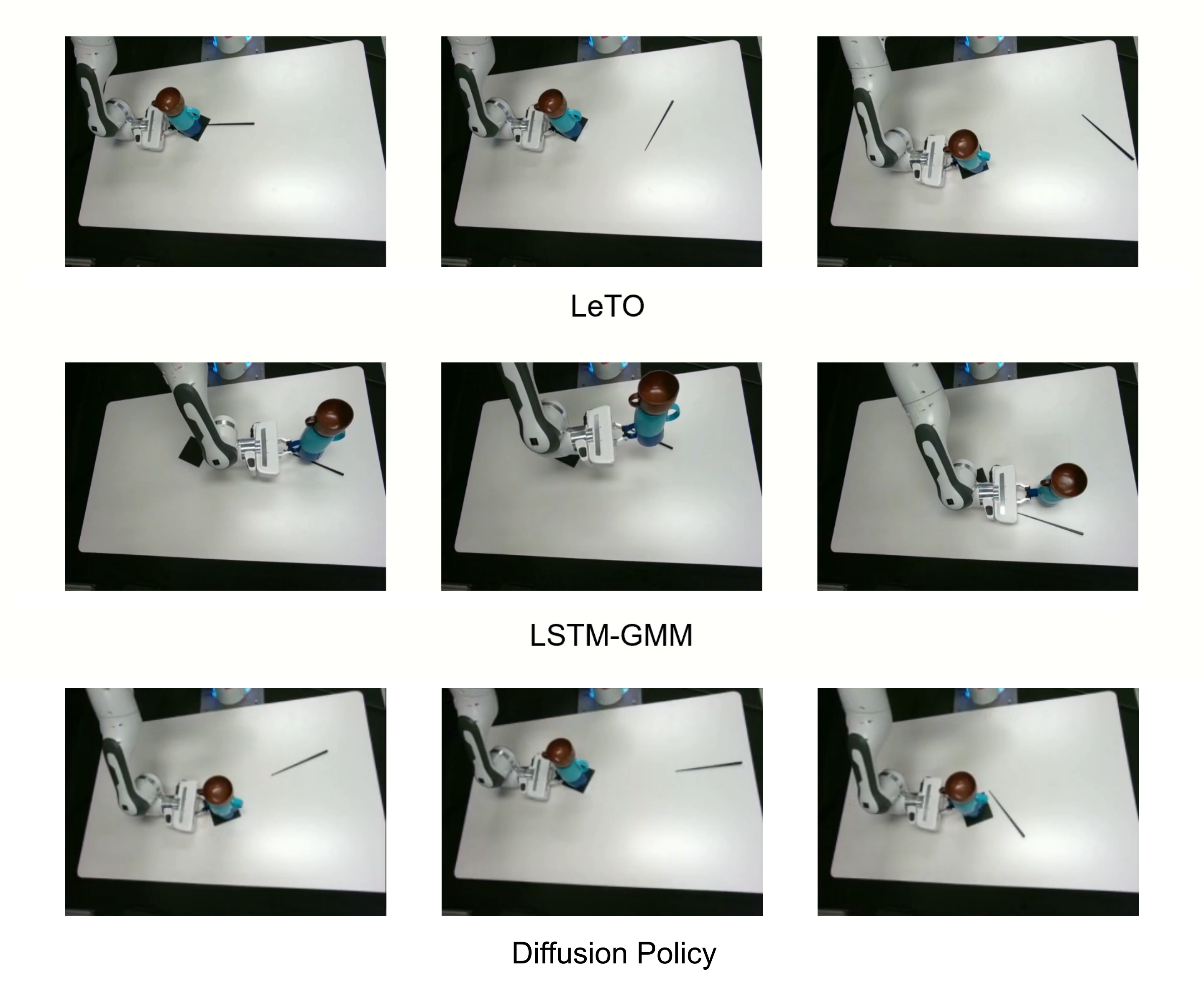}
\end{overpic}
\caption{When chopsticks fall, which is a type of OOD data, both the diffusion policy and LeTO can still move the stacked cups and place them on the black board. In contrast, LSTM-GMM often exhibits erroneous behaviors, such as moving to incorrect locations or acting erratically. }
\label{fig:ood}
\end{figure}

The objective of the ``Move-the-stack" task is that the robot grasps a set of stacked objects, smoothly transports them, and ultimately places them onto a black board, ensuring none of the stacked objects fall off. The initial positions of the robot and the black board are randomized. The policy inputs end-effector positions and images from a third-person perspective camera and controls the end-effector position. We introduce the velocity constraint [-1,1] and the acceleration constraint [-0.25,0.25] for LeTO. In particular, while we did not place chopsticks on top during human demonstration collection, a chopstick was added to the stacked cups during policy rollout. 

For imitation learning algorithms, the task is challenging because: 1) The policy needs to generate very smooth trajectories. Any roughness in the trajectory can lead to vibrations at the robot's end-effector, causing the cups to shake. This vibration can accumulate and cause the chopstick to drop. 2) Since a chopstick is introduced during policy rollout, this out-of-distribution (OOD) observation adds greater uncertainty to the network input. This uncertainty may make the network's output jerkier, resulting in a poorer quality trajectory.

\begin{table}[t]
\caption{Metrics for evaluating the Move-the-stack task. ``acc-peak" represents the highest acceleration achieved in any of the test trials while ``avg-acc-mean/max/std" are the average of mean/max/std values across all test trials.}
\centering
\begin{tabular}{|c|c|c|c|}
\hline 
& avg-acc-{mean}/{max}/{std} & succ-rate & acc-peak \\
\hline 
LSTM-GMM  &{0.053}/{0.480}/{0.081}& 4/20 & 0.647\\
DiffusionPolicy  &\textbf{{0.037}}/{0.336}/{0.055}& 4/20 & 0.721\\
LeTO & \textbf{{0.037}/{0.219}/{0.042}} & \textbf{13/20} & \textbf{0.250} \\
\hline
\end{tabular}
\label{tb:move}
\end{table}

\begin{table}[t]
\caption{Metrics for evaluating the Arrange-chopsticks task.
``contact-peak" represents the highest contact force achieved in any of the test trials while ``avg-contact-max" is the average of max contact forces across all test trials.}
\centering
\begin{tabular}{|c|c|c|c|}
\hline 
& avg-contact-max & contact-peak & succ-rate \\
\hline 
LSTM-GMM  &13.2~N & 27.3~N& 0/20\\
DiffusionPolicy  &13.9~N &24.8~N & \textbf{18/20}\\
LeTO &\textbf{11.7~N}& \textbf{16.1~N} & \textbf{17/20} \\
\hline
\end{tabular}
\label{tb:arrange}
\end{table}

The results are shown in Table~\ref{tb:move}. From the results, LeTO excels in terms of trajectory smoothness and, correspondingly, also boasts the highest success rate. This shows that due to the presence of end-to-end trajectory optimization, LeTO can robustly maintain its effectiveness even when facing OOD data. Another point worth noting is that when chopsticks fall (which is also a type of OOD data), both the diffusion policy and LeTO can still move the stacked cups and place them on the black board. In contrast, LSTM-GMM often exhibits erroneous behaviors, such as moving to incorrect locations or acting erratically (see Fig.~\ref{fig:ood} and the attached video). 

{The Move-the-stack task highlights the importance of trajectory optimization. For methods solely based on neural networks, there is a lack of control over the generated trajectories, resulting in trajectories that are less smooth and more uncertain. The inherent uninterpretability of neural networks introduces greater uncertainty.}

\subsection{Arrange-chopsticks}
The objective of the ``Arrange-chopsticks" task is to pick up two randomly placed chopsticks and arrange them neatly on a bowl. The policy inputs end-effector poses and images from two cameras, a third-person perspective camera and a wrist-mounted camera, and controls the end-effector position, yaw angle, and grasping. 

The challenges of this task include: 1. It is a relatively long-range task that requires the policy to have the capability to represent complex tasks. 2. Due to the slender nature of chopsticks, there is a high risk of the robot making forceful contact with the table when attempting to grasp them. For the sake of hardware safety, such collisions should be minimized as much as possible.

We introduce both position, velocity, and acceleration constraints for LeTO. During the demonstration, the lowest position of the robot's end-effector on the z-axis was recorded at 0.0501~m. We have opted for 0.0475~m as the z-axis positional constraint within the LeTO framework. This choice is made with the task's completion in mind, where the goal is not to prevent any contact between the robot's end-effector and the table surface, but rather to allow for the slightest possible contact that still enables the picking up of slender chopsticks.

The results are shown in Table~\ref{tb:arrange}. Both LeTO and diffusion policy have high success rate, but the contact forces in LeTO tests are significantly smaller. {Collisions with the environment are quite common for manipulation tasks like picking up very fine chopsticks from the tabletop. One can imagine that if the table were made of glass, such collisions could be very hazardous. Moreover, consider a scenario in which a robot frequently collides with its environment without constraints. Even if this does not affect performance in the short term, it could significantly reduce the robot’s lifespan over time.}

{The reduction in contact force with the tabletop is a result of the implementation of the LeTO differentiable optimization layer, which incorporates position constraints that allow for trajectory optimization. In contrast, existing "black box" approaches, which are solely based on neural networks, do not ensure outputs that consistently meet these constraints, leading to unpredictable and potentially large impact forces with the tabletop.}

{Given that robots physically interact with their environment and humans, factors like safety, interpretability, stability, and robustness are crucial. Fully neural network-based approaches pose challenges for real-world deployment due to their lack of interpretability. For instance, even if the algorithm performs well under most conditions, the absence of constraints means that significant deviations in a few edge cases can lead to severe problems due to the physical nature of robots.}

\section{Discussion and Future Work}\label{sec:diss}
In this paper, we present LeTO, a framework for learning
constrained visuomotor policy with differentiable trajectory
optimization. By balancing the objective of minimizing errors
with the need to satisfy trajectory constraints, LeTO allows
for deployment in tasks where safety and reliability are paramount. 

\subsection{Training Time and Training Stability}

\begin{table}[t]
\caption{{The training time for 20 epochs on the square ph task with a batch size of 64, conducted on a single GeForce RTX 3080. The hyperparameters for diffusion policy and LSTM-GMM are adopted from the default parameters reported in \cite{mandlekar2022matters,chi2023diffusionpolicy}, respectively.}}
\centering
{
\begin{tabular}{|c|c|c|c|}
\hline
\text{Policy} & \text{LSTM-GMM} & \text{DiffusionPolicy} & \text{LeTO} \\
\hline
\text{Training Time} & \text{2.90 hours} & \text{0.92 hours} & \text{7.25 hours} \\
\hline
\end{tabular}}
\label{tb:time}
\end{table}

As shown in Remark~\ref{rm:1}, the optimization problem of the DTO layer is always feasible during training, given that the positions (and orientations) in the dataset meet the set position constraints. The consistent feasibility of the DTO layer ensures training stability, which is evident from the training logs of LeTO available in our open-source code.

Table~\ref{tb:time} compares the training times of LSTM-GMM, diffusion policy, and LeTO. We acknowledge that one limitation of LeTO is the computational overhead introduced by the differentiable optimization layer. Future work could look into improving the computational efficiency of the differentiable optimization process, potentially through the use of more efficient solvers or by learning approximations of the optimization layer.

\subsection{LeTO in Low-data and High-data Regimes}

We would like to highlight that networks with the training objective of minimizing errors to fit human demonstrated actions can easily incorporate our proposed DTO layer at the head. This suggests that for scalability, one could consider designing a LeTO variant with transformer, enhancing the model's capability to scale to large datasets. {However, the time and cost involved in training could still be a limiting factor. As shown in Table~\ref{tb:time}, LeTO exhibits a significantly slower training compared to other methods. As highlighted in \cite{amos2017optnet}, solving optimization problems with precision, as done in this case, demonstrates cubic complexity relative to the number of variables and/or constraints. Therefore, future research focused on accelerating differentiable optimization is pivotal for scalability. }

Furthermore, exploring policies that excel in low-data regimes based on differentiable optimization layers is an intriguing research direction. By constructing differentiable optimization layers using prior knowledge of the model, policy may reduce reliance on data.

\subsection{LeTO in Reinforcement Learning}

Due to the differentiability, the DTO layer can be designed to integrate into the robot learning pipelines, enabling end-to-end optimization and inference. This opens up the possibility of extending LeTO to reinforcement learning, offering a promising future research direction. By integrating differentiable trajectory optimization with reinforcement learning, it is possible to ensure that the generated actions comply with the designed constraints, thus significantly improving the safety and interpretability of the policy. 

\subsection{LeTO in High-safety Scenarios}

The safe and constraint-controlled output form of LeTO holds significant potential in robot learning scenarios where safety is of utmost importance, such as in human-robot interaction and surgical robotics. Investigating the application of LeTO, or more broadly, robot policies that integrate differentiable optimization in these settings, represents a highly interesting and promising research direction. 

{Additionally, since imitation learning provides a high-level controller responsible for decision-making and high-level action generation, LeTO's trajectory optimization capabilities enable it to generate reference motions that are dynamically feasible for a low-level controller. In scenarios requiring high dynamics or heavy loads for manipulation, LeTO may have certain advantages and safety assurances when combined with more specialized lower-level controllers to accomplish these tasks.}

\subsection{Complex Obstacle Avoidance and Nonconvex Optimization}

{As stated in our paper, LeTO's constraints must be convex. In complex environments filled with obstacles, these constraints become highly complicated and nonconvex. Therefore, LeTO does not have the ability to generate collision-free actions in confined environments filled with diverse and complex obstacles. We believe that developing an imitation learning algorithm capable of achieving collision-free performance in complex environments while executing dexterous manipulation tasks will be a very promising and important research direction.}

\section{acknowledgements}
This work was partially supported by the United States Department of Agriculture (USDA; No. 2023-67021-39072 and 2024-67021-42878) as well as National Science Foundation (NSF; No. 2423068). This article solely reflects the opinions and conclusions of its authors and not USDA or NSF.

\ifCLASSOPTIONcaptionsoff
  \newpage
\fi

\bibliographystyle{IEEEtran}
\bibliography{IEEEabrv,paperref}

\begin{IEEEbiography}[{\includegraphics[width=1in,height=1.25in,clip,keepaspectratio]{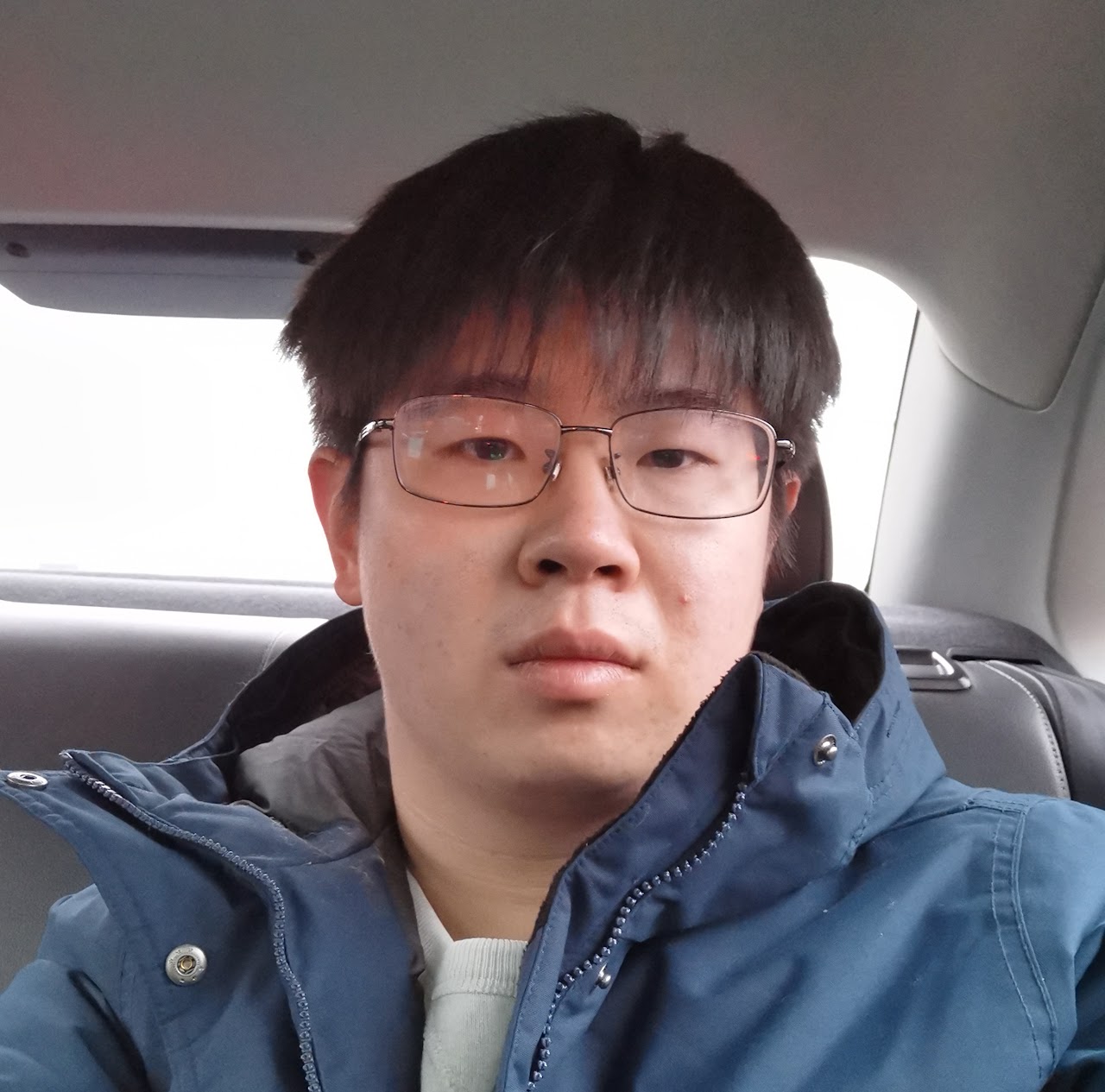}}]{Zhengtong Xu} received his Bachelor of Engineering degree in mechanical engineering from Huazhong University of Science and Technology, China in 2020. He is currently pursuing his Ph.D. at Purdue University. 

His research focuses on robot learning.
\end{IEEEbiography}
\vskip -2\baselineskip plus -1fil
\begin{IEEEbiography}[{\includegraphics[width=1in,height=1.25in,clip,keepaspectratio]{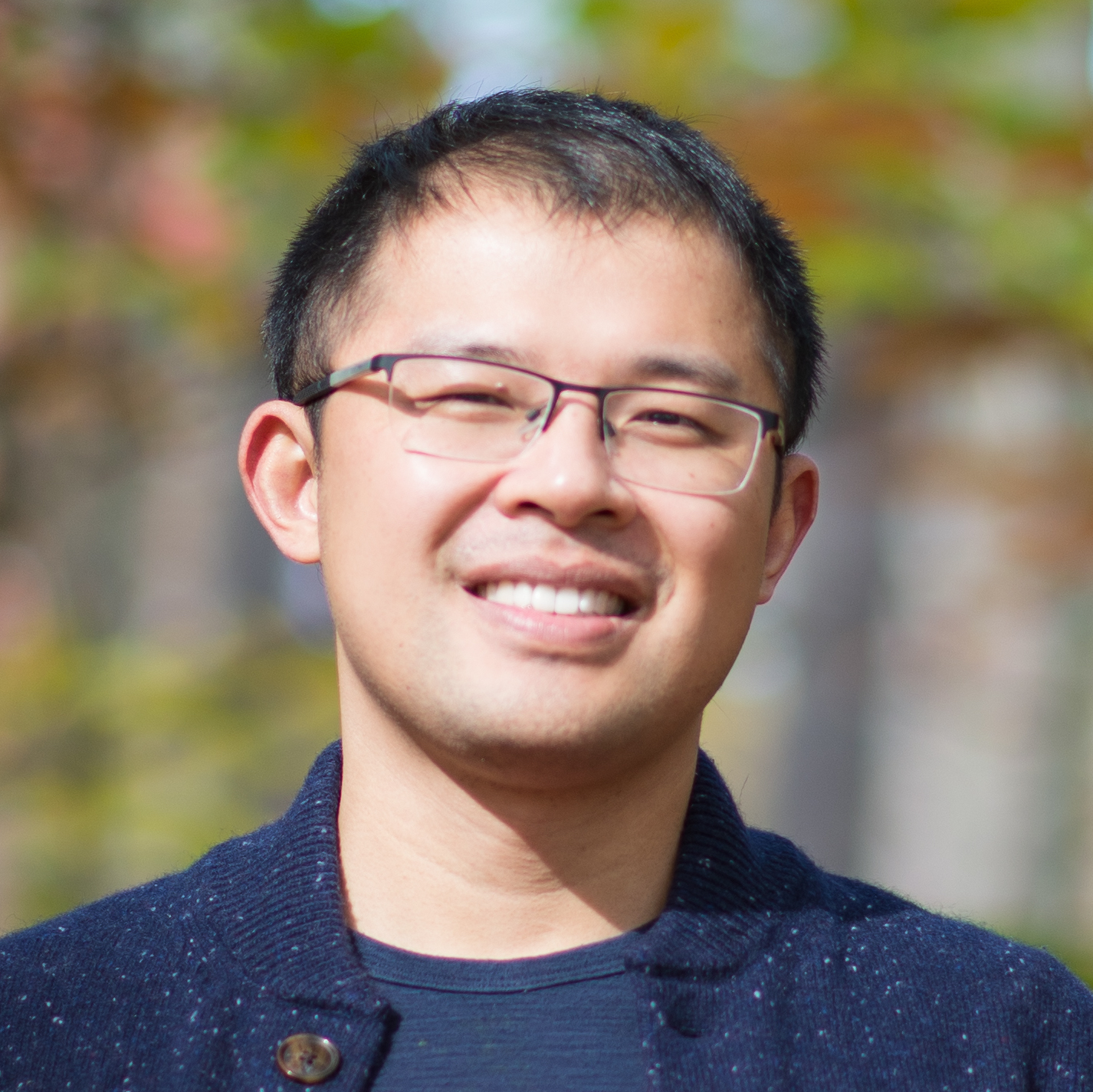}}]{Yu She} is an assistant professor at Purdue University Edwardson School of Industrial Engineering. Prior to that, he was a postdoctoral researcher in the Computer Science and Artificial Intelligence Laboratory at MIT from 2018 to 2021. He earned his Ph.D. degree in the Department of Mechanical Engineering at the Ohio State University in 2018. His research, at the intersection of mechanical design, sensory perception, and dynamic control, explores human-safe collaborative robots, soft robotics, and robotic manipulation.
\end{IEEEbiography}

\end{document}

%% file: before_document.tex
\usepackage{graphicx}
\usepackage[percent]{overpic}
\usepackage{subfig}
\usepackage{amsmath}
\usepackage{amsthm} 
\usepackage{amssymb}
\usepackage[font=footnotesize]{caption} 
\usepackage{subfig} 
\usepackage[noadjust]{cite} 
\usepackage{color}
\usepackage{algorithm} 
\usepackage{algpseudocode} 
\usepackage{enumerate}
\usepackage {hyperref} 
\hypersetup {colorlinks,allcolors=black}
\usepackage[left=0.75in, right=0.75in, top=0.75in, bottom=0.75in]{geometry}
\usepackage{authblk} 
\usepackage{arydshln} 

\usepackage{bm}
\usepackage{tikz}
\usetikzlibrary{calc} 
\usetikzlibrary{shapes} 
\usetikzlibrary{chains}
\usetikzlibrary{fit}
\usetikzlibrary{arrows}
\usetikzlibrary{decorations.text} 
\usetikzlibrary{decorations.markings}
\usetikzlibrary{decorations.pathmorphing} 
\usetikzlibrary{shadows}
\usetikzlibrary{patterns}
\usetikzlibrary{matrix}
\usepackage{pgfplots}
\usepackage[europeanresistors]{circuitikz}
\usepackage[outline]{contour} 
\contourlength{1.5pt}

\newtheorem{remark}{Remark}

\graphicspath{{figures/}}

%% file: LeTO.bbl
\begin{thebibliography}{10}
\providecommand{\url}[1]{#1}
\csname url@samestyle\endcsname
\providecommand{\newblock}{\relax}
\providecommand{\bibinfo}[2]{#2}
\providecommand{\BIBentrySTDinterwordspacing}{\spaceskip=0pt\relax}
\providecommand{\BIBentryALTinterwordstretchfactor}{4}
\providecommand{\BIBentryALTinterwordspacing}{\spaceskip=\fontdimen2\font plus
\BIBentryALTinterwordstretchfactor\fontdimen3\font minus \fontdimen4\font\relax}
\providecommand{\BIBforeignlanguage}[2]{{%
\expandafter\ifx\csname l@#1\endcsname\relax
\typeout{** WARNING: IEEEtran.bst: No hyphenation pattern has been}%
\typeout{** loaded for the language `#1'. Using the pattern for}%
\typeout{** the default language instead.}%
\else
\language=\csname l@#1\endcsname
\fi
#2}}
\providecommand{\BIBdecl}{\relax}
\BIBdecl

\bibitem{pomerleau1988alvinn}
D.~A. Pomerleau, ``Alvinn: An autonomous land vehicle in a neural network,'' \emph{Advances in neural information processing systems}, vol.~1, 1988.

\bibitem{chi2024universal}
C.~Chi, Z.~Xu, C.~Pan, E.~Cousineau, B.~Burchfiel, S.~Feng, R.~Tedrake, and S.~Song, ``Universal manipulation interface: In-the-wild robot teaching without in-the-wild robots,'' \emph{arXiv preprint arXiv:2402.10329}, 2024.

\bibitem{xu2024unit}
\BIBentryALTinterwordspacing
Z.~Xu, R.~Uppuluri, X.~Zhang, C.~Fitch, P.~G. Crandall, W.~Shou, D.~Wang, and Y.~She, ``{UniT}: Unified tactile representation for robot learning,'' 2024. [Online]. Available: \url{https://arxiv.org/abs/2408.06481}
\BIBentrySTDinterwordspacing

\bibitem{fu2024mobile}
Z.~Fu, T.~Z. Zhao, and C.~Finn, ``Mobile aloha: Learning bimanual mobile manipulation with low-cost whole-body teleoperation,'' \emph{arXiv preprint arXiv:2401.02117}, 2024.

\bibitem{zhu2022viola}
Y.~Zhu, A.~Joshi, P.~Stone, and Y.~Zhu, ``Viola: Imitation learning for vision-based manipulation with object proposal priors,'' \emph{arXiv preprint arXiv:2210.11339}, 2022.

\bibitem{wang2024poco}
L.~Wang, J.~Zhao, Y.~Du, E.~H. Adelson, and R.~Tedrake, ``Poco: Policy composition from and for heterogeneous robot learning,'' \emph{arXiv preprint arXiv:2402.02511}, 2024.

\bibitem{xu2023gan}
X.~Xu, M.~You, H.~Zhou, Z.~Qian, W.~Xu, and B.~He, ``Gan-based editable movement primitive from high-variance demonstrations,'' \emph{IEEE Robotics and Automation Letters}, vol.~8, no.~8, pp. 4593--4600, 2023.

\bibitem{pignat2021generative}
E.~Pignat, H.~Girgin, and S.~Calinon, ``Generative adversarial training of product of policies for robust and adaptive movement primitives,'' in \emph{Conference on Robot Learning}.\hskip 1em plus 0.5em minus 0.4em\relax PMLR, 2021, pp. 1456--1470.

\bibitem{bahl2022human}
S.~Bahl, A.~Gupta, and D.~Pathak, ``Human-to-robot imitation in the wild,'' \emph{arXiv preprint arXiv:2207.09450}, 2022.

\bibitem{wang2023mimicplay}
C.~Wang, L.~Fan, J.~Sun, R.~Zhang, L.~Fei-Fei, D.~Xu, Y.~Zhu, and A.~Anandkumar, ``Mimicplay: Long-horizon imitation learning by watching human play,'' \emph{arXiv preprint arXiv:2302.12422}, 2023.

\bibitem{qian2023robot}
Z.~Qian, M.~You, H.~Zhou, X.~Xu, and B.~He, ``Robot learning from human demonstrations with inconsistent contexts,'' \emph{Robotics and Autonomous Systems}, vol. 166, p. 104466, 2023.

\bibitem{zhao2023learning}
T.~Z. Zhao, V.~Kumar, S.~Levine, and C.~Finn, ``Learning fine-grained bimanual manipulation with low-cost hardware,'' \emph{arXiv preprint arXiv:2304.13705}, 2023.

\bibitem{jarrett2020strictly}
D.~Jarrett, I.~Bica, and M.~van~der Schaar, ``Strictly batch imitation learning by energy-based distribution matching,'' \emph{Advances in Neural Information Processing Systems}, vol.~33, pp. 7354--7365, 2020.

\bibitem{florence2022implicit}
P.~Florence, C.~Lynch, A.~Zeng, O.~A. Ramirez, A.~Wahid, L.~Downs, A.~Wong, J.~Lee, I.~Mordatch, and J.~Tompson, ``Implicit behavioral cloning,'' in \emph{Conference on Robot Learning}.\hskip 1em plus 0.5em minus 0.4em\relax PMLR, 2022, pp. 158--168.

\bibitem{chi2023diffusionpolicy}
C.~Chi, S.~Feng, Y.~Du, Z.~Xu, E.~Cousineau, B.~Burchfiel, and S.~Song, ``Diffusion policy: Visuomotor policy learning via action diffusion,'' in \emph{Proceedings of Robotics: Science and Systems (RSS)}, 2023.

\bibitem{zucker2013chomp}
M.~Zucker, N.~Ratliff, A.~D. Dragan, M.~Pivtoraiko, M.~Klingensmith, C.~M. Dellin, J.~A. Bagnell, and S.~S. Srinivasa, ``Chomp: Covariant hamiltonian optimization for motion planning,'' \emph{The International journal of robotics research}, vol.~32, no. 9-10, pp. 1164--1193, 2013.

\bibitem{schulman2014motion}
J.~Schulman, Y.~Duan, J.~Ho, A.~Lee, I.~Awwal, H.~Bradlow, J.~Pan, S.~Patil, K.~Goldberg, and P.~Abbeel, ``Motion planning with sequential convex optimization and convex collision checking,'' \emph{The International Journal of Robotics Research}, vol.~33, no.~9, pp. 1251--1270, 2014.

\bibitem{mellinger2011minimum}
D.~Mellinger and V.~Kumar, ``Minimum snap trajectory generation and control for quadrotors,'' in \emph{2011 IEEE international conference on robotics and automation}.\hskip 1em plus 0.5em minus 0.4em\relax IEEE, 2011, pp. 2520--2525.

\bibitem{zhang2020optimization}
X.~Zhang, A.~Liniger, and F.~Borrelli, ``Optimization-based collision avoidance,'' \emph{IEEE Transactions on Control Systems Technology}, vol.~29, no.~3, pp. 972--983, 2020.

\bibitem{zhang2018deep}
T.~Zhang, Z.~McCarthy, O.~Jow, D.~Lee, X.~Chen, K.~Goldberg, and P.~Abbeel, ``Deep imitation learning for complex manipulation tasks from virtual reality teleoperation,'' in \emph{2018 IEEE International Conference on Robotics and Automation (ICRA)}.\hskip 1em plus 0.5em minus 0.4em\relax IEEE, 2018, pp. 5628--5635.

\bibitem{florence2019self}
P.~Florence, L.~Manuelli, and R.~Tedrake, ``Self-supervised correspondence in visuomotor policy learning,'' \emph{IEEE Robotics and Automation Letters}, vol.~5, no.~2, pp. 492--499, 2019.

\bibitem{bojarski2016end}
M.~Bojarski, D.~Del~Testa, D.~Dworakowski, B.~Firner, B.~Flepp, P.~Goyal, L.~D. Jackel, M.~Monfort, U.~Muller, J.~Zhang \emph{et~al.}, ``End to end learning for self-driving cars,'' \emph{arXiv preprint arXiv:1604.07316}, 2016.

\bibitem{ross2011reduction}
S.~Ross, G.~Gordon, and D.~Bagnell, ``A reduction of imitation learning and structured prediction to no-regret online learning,'' in \emph{Proceedings of the fourteenth international conference on artificial intelligence and statistics}.\hskip 1em plus 0.5em minus 0.4em\relax JMLR Workshop and Conference Proceedings, 2011, pp. 627--635.

\bibitem{rahmatizadeh2018vision}
R.~Rahmatizadeh, P.~Abolghasemi, L.~B{\"o}l{\"o}ni, and S.~Levine, ``Vision-based multi-task manipulation for inexpensive robots using end-to-end learning from demonstration,'' in \emph{2018 IEEE international conference on robotics and automation (ICRA)}.\hskip 1em plus 0.5em minus 0.4em\relax IEEE, 2018, pp. 3758--3765.

\bibitem{wu2020spatial}
J.~Wu, X.~Sun, A.~Zeng, S.~Song, J.~Lee, S.~Rusinkiewicz, and T.~Funkhouser, ``Spatial action maps for mobile manipulation,'' \emph{arXiv preprint arXiv:2004.09141}, 2020.

\bibitem{zeng2021transporter}
A.~Zeng, P.~Florence, J.~Tompson, S.~Welker, J.~Chien, M.~Attarian, T.~Armstrong, I.~Krasin, D.~Duong, V.~Sindhwani \emph{et~al.}, ``Transporter networks: Rearranging the visual world for robotic manipulation,'' in \emph{Conference on Robot Learning}.\hskip 1em plus 0.5em minus 0.4em\relax PMLR, 2021, pp. 726--747.

\bibitem{avigal2022speedfolding}
Y.~Avigal, L.~Berscheid, T.~Asfour, T.~Kr{\"o}ger, and K.~Goldberg, ``Speedfolding: Learning efficient bimanual folding of garments,'' in \emph{2022 IEEE/RSJ International Conference on Intelligent Robots and Systems (IROS)}.\hskip 1em plus 0.5em minus 0.4em\relax IEEE, 2022, pp. 1--8.

\bibitem{shafiullah2022behavior}
N.~M. Shafiullah, Z.~Cui, A.~A. Altanzaya, and L.~Pinto, ``Behavior transformers: Cloning $ k $ modes with one stone,'' \emph{Advances in neural information processing systems}, vol.~35, pp. 22\,955--22\,968, 2022.

\bibitem{mandlekar2022matters}
A.~Mandlekar, D.~Xu, J.~Wong, S.~Nasiriany, C.~Wang, R.~Kulkarni, L.~Fei-Fei, S.~Savarese, Y.~Zhu, and R.~Mart{\'\i}n-Mart{\'\i}n, ``What matters in learning from offline human demonstrations for robot manipulation,'' in \emph{Proc. Conf. Robot Learn.}, 2022, pp. 1678--1690.

\bibitem{Pfrommer2020}
\BIBentryALTinterwordspacing
S.~Pfrommer, M.~Halm, and M.~Posa, ``{ContactNets: Learning Discontinuous Contact Dynamics with Smooth, Implicit Representations},'' in \emph{The Conference on Robot Learning (CoRL)}, 2020. [Online]. Available: \url{https://proceedings.mlr.press/v155/pfrommer21a.html}
\BIBentrySTDinterwordspacing

\bibitem{Bianchini2022}
\BIBentryALTinterwordspacing
B.~Bianchini, M.~Halm, N.~Matni, and M.~Posa, ``Generalization bounded implicit learning of nearly discontinuous functions,'' in \emph{Proceedings of The 4th Annual Learning for Dynamics and Control Conference (L4DC)}, ser. Proceedings of Machine Learning Research, R.~Firoozi, N.~Mehr, E.~Yel, R.~Antonova, J.~Bohg, M.~Schwager, and M.~Kochenderfer, Eds., vol. 168.\hskip 1em plus 0.5em minus 0.4em\relax PMLR, 23--24 Jun 2022, pp. 1112--1124. [Online]. Available: \url{https://proceedings.mlr.press/v168/bianchini22a.html}
\BIBentrySTDinterwordspacing

\bibitem{amos2018differentiable}
B.~Amos, I.~Jimenez, J.~Sacks, B.~Boots, and J.~Z. Kolter, ``Differentiable mpc for end-to-end planning and control,'' \emph{Advances in neural information processing systems}, vol.~31, 2018.

\bibitem{retchin2023koopman}
\BIBentryALTinterwordspacing
M.~Retchin, B.~Amos, S.~Brunton, and S.~Song, ``Koopman constrained policy optimization: A koopman operator theoretic method for differentiable optimal control in robotics,'' in \emph{ICML 2023 Workshop on Differentiable Almost Everything: Differentiable Relaxations, Algorithms, Operators, and Simulators}, 2023. [Online]. Available: \url{https://openreview.net/forum?id=3W7vPqWCeM}
\BIBentrySTDinterwordspacing

\bibitem{xu2024letac}
Z.~Xu and Y.~She, ``{LeTac-MPC}: Learning model predictive control for tactile-reactive grasping,'' \emph{IEEE Transactions on Robotics}, vol.~40, pp. 4376--4395, 2024.

\bibitem{xiao2021barriernet}
W.~Xiao, R.~Hasani, X.~Li, and D.~Rus, ``Barriernet: A safety-guaranteed layer for neural networks,'' \emph{arXiv preprint arXiv:2111.11277}, 2021.

\bibitem{diehl2022differentiable}
C.~Diehl, J.~Adamek, M.~Kr{\"u}ger, F.~Hoffmann, and T.~Bertram, ``Differentiable constrained imitation learning for robot motion planning and control,'' \emph{arXiv preprint arXiv:2210.11796}, 2022.

\bibitem{karkus2023diffstack}
P.~Karkus, B.~Ivanovic, S.~Mannor, and M.~Pavone, ``Diffstack: A differentiable and modular control stack for autonomous vehicles,'' in \emph{Conference on Robot Learning}.\hskip 1em plus 0.5em minus 0.4em\relax PMLR, 2023, pp. 2170--2180.

\bibitem{wan2024difftop}
W.~Wan, Y.~Wang, Z.~Erickson, and D.~Held, ``Difftop: Differentiable trajectory optimization for deep reinforcement and imitation learning,'' \emph{arXiv preprint arXiv:2402.05421}, 2024.

\bibitem{ratliff2018riemannian}
N.~D. Ratliff, J.~Issac, D.~Kappler, S.~Birchfield, and D.~Fox, ``Riemannian motion policies,'' \emph{arXiv preprint arXiv:1801.02854}, 2018.

\bibitem{li2021rmp2}
A.~Li, C.-A. Cheng, M.~A. Rana, M.~Xie, K.~Van~Wyk, N.~Ratliff, and B.~Boots, ``Rmp2: A structured composable policy class for robot learning,'' \emph{arXiv preprint arXiv:2103.05922}, 2021.

\bibitem{cheng2020rmp}
C.-A. Cheng, M.~Mukadam, J.~Issac, S.~Birchfield, D.~Fox, B.~Boots, and N.~Ratliff, ``Rmp flow: A computational graph for automatic motion policy generation,'' in \emph{Algorithmic Foundations of Robotics XIII: Proceedings of the 13th Workshop on the Algorithmic Foundations of Robotics 13}.\hskip 1em plus 0.5em minus 0.4em\relax Springer, 2020, pp. 441--457.

\bibitem{mandlekar2020learning}
A.~Mandlekar, D.~Xu, R.~Mart{\'\i}n-Mart{\'\i}n, S.~Savarese, and L.~Fei-Fei, ``Learning to generalize across long-horizon tasks from human demonstrations,'' \emph{arXiv preprint arXiv:2003.06085}, 2020.

\bibitem{mandlekar2020iris}
A.~Mandlekar, F.~Ramos, B.~Boots, S.~Savarese, L.~Fei-Fei, A.~Garg, and D.~Fox, ``Iris: Implicit reinforcement without interaction at scale for learning control from offline robot manipulation data,'' in \emph{2020 IEEE International Conference on Robotics and Automation (ICRA)}.\hskip 1em plus 0.5em minus 0.4em\relax IEEE, 2020, pp. 4414--4420.

\bibitem{amos2017optnet}
B.~Amos and J.~Z. Kolter, ``{OptNet}: Differentiable optimization as a layer in neural networks,'' in \emph{Proc. 34th Int. Conf. Mach. Learn.}, 2017, pp. 136--145.

\bibitem{boyd2004convex}
S.~Boyd, S.~P. Boyd, and L.~Vandenberghe, \emph{Convex optimization}.\hskip 1em plus 0.5em minus 0.4em\relax Cambridge university press, 2004.

\bibitem{liu2022robot}
H.~Liu, S.~Nasiriany, L.~Zhang, Z.~Bao, and Y.~Zhu, ``Robot learning on the job: Human-in-the-loop autonomy and learning during deployment,'' in \emph{Robotics: Science and Systems (RSS)}, 2023.

\end{thebibliography}
